\documentclass[10pt,journal,compsoc]{IEEEtran}
\ifCLASSOPTIONcompsoc
  \usepackage[nocompress]{cite}
\else
  \usepackage{cite}
\fi
\ifCLASSINFOpdf
\else
\fi
\usepackage{bm}
\usepackage{amssymb}
\usepackage{latexsym}
\usepackage{dsfont}
\usepackage{multirow}
\usepackage{amsfonts}
\usepackage{amsmath}
\usepackage{algorithm}
\usepackage{algorithmic}
\usepackage{graphicx}
\usepackage{hyperref}
\usepackage{color}
\usepackage{soul} 

\DeclareMathAlphabet{\mathsfit}{\encodingdefault}{\sfdefault}{m}{sl}
\SetMathAlphabet{\mathsfit}{bold}{\encodingdefault}{\sfdefault}{bx}{n}

\def\sY{{\mathbb{Y}}}

\newcommand{\red}[1] {\textcolor{black}{{#1}}}

\def\ie{{\it i.e.}}
\def\eg{{\it e.g.}}
\def\et{{\it et al.}}
\def\etc{{\it etc.}}

\begin{document}
%
\title{Imperceptible Transfer Attack and Defense on 3D Point Cloud Classification}

\author{Daizong~Liu,~\IEEEmembership{Student~Member,~IEEE,}
        and~Wei~Hu,~\IEEEmembership{Senior~Member,~IEEE}
\IEEEcompsocitemizethanks{\IEEEcompsocthanksitem D. Liu and W. Hu are with Wangxuan Institute of Computer Technology, Peking University, No. 128, Zhongguancun North Street, Beijing, China. E-mail: dzliu@stu.pku.edu.cn, forhuwei@pku.edu.cn. 
\IEEEcompsocthanksitem Corresponding author: Wei Hu.}}

%
%

\markboth{Journal of \LaTeX\ Class Files,~Vol.~14, No.~8, August~2015}%
{Shell \MakeLowercase{\textit{et al.}}: Bare Demo of IEEEtran.cls for Computer Society Journals}
%



\IEEEtitleabstractindextext{%
\begin{abstract}
Although many efforts have been made into attack and defense on the 2D image domain in recent years, few methods explore the vulnerability of 3D models. Existing 3D attackers generally perform point-wise perturbation over point clouds, resulting in deformed structures or outliers, which is easily perceivable by humans. Moreover, their adversarial examples are generated under the white-box setting, which frequently suffers from low success rates when transferred to attack remote black-box models. In this paper, we study 3D point cloud attacks from two new and challenging perspectives by proposing a novel Imperceptible Transfer Attack (ITA): 1) Imperceptibility: we constrain the perturbation direction of each point along its normal vector of the neighborhood surface, leading to generated examples with similar geometric properties and thus enhancing the imperceptibility. 2) Transferability: we develop an adversarial transformation model to generate the most harmful distortions and enforce the adversarial examples to resist it, improving their transferability to unknown black-box models. Further, we propose to train more robust black-box 3D models to defend against such ITA attacks by learning more discriminative point cloud representations. Extensive evaluations demonstrate that our ITA attack is more imperceptible and transferable than state-of-the-arts and validate the superiority of our defense strategy.
\end{abstract}

}

\maketitle

\IEEEdisplaynontitleabstractindextext

%
\IEEEpeerreviewmaketitle

\IEEEraisesectionheading{\section{Introduction}
\label{sec:introduction}}

\IEEEPARstart{D}{eep} Neural Networks (DNNs) are known to be vulnerable to adversarial examples \cite{szegedy2013intriguing,goodfellow2014explaining}, which are indistinguishable from legitimate ones by adding trivial perturbations but often lead to incorrect model prediction.
Many efforts have been made into attacks on the 2D image field \cite{dong2018boosting,madry2017towards,kurakin2016adversarial,tu2019autozoom}, which often add pixel-wise noise on images to deceive DNNs in the testing stage. 
Nevertheless, adversarial attacks on 3D point clouds---discrete representations of 3D scenes or objects that consist of a set of points residing on irregular domains---are still relatively under-explored. 
Point cloud attacks are crucial in various safety-critical
applications such as autonomous driving \cite{chen2017multi,yue2018lidar} and robotic grasping \cite{varley2017shape,zhong2020reliable}, and face several challenges in real scenarios. 

\begin{figure*}[t!]
\centering
\includegraphics[width=0.95\textwidth]{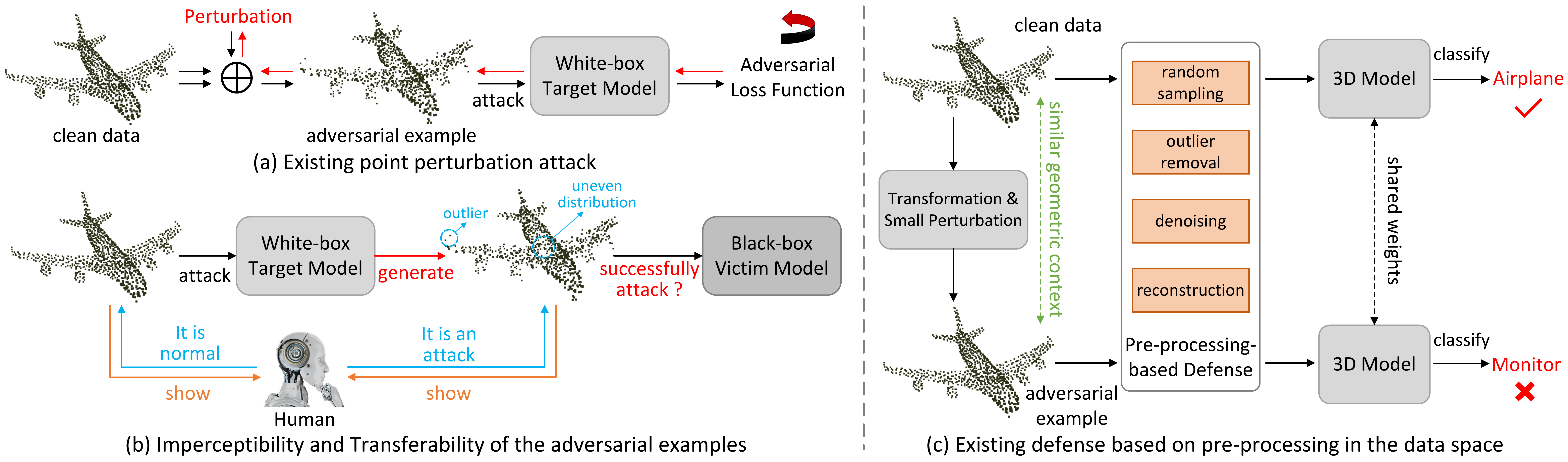}
\caption{(a) Existing point perturbation attack is under the white-box setting, which shifts each point along the xyz directions with the supervision of an adversarial loss function. (b) Perturbed point clouds generated by existing white-box attacks generally exhibit outliers or uneven distribution in local regions, which are perceptible to humans. Besides, they tend to overfit the target model and have low success rates when transferred to attack a new black-box victim model.
(c) Existing defense methods generally adopt pre-processing to restore the adversarial point clouds in the data space, which fails to capture high-level representations in the latent space that are crucial to defense.
}
\label{fig:introduction}
\end{figure*}

Given a point cloud, traditional works \cite{zhang2019adversarial,xiang2019generating} mostly perform adversarial attacks by adding or deleting a small set of points, which enforces the point cloud to be misclassified by a 3D model. These works generally utilize gradient-guided attack methods to explore the best scheme of point modification, addition, and deletion. 
Recently, as demonstrated in Figure \ref{fig:introduction}(a), more attempts \cite{zhang2019defense,liu2019extending,tsai2020robust,wen2020geometry} focus on the point-wise perturbation attack by changing coordinates of the point cloud in a learnable way with the Fast Gradient Sign Method (FGSM) \cite{goodfellow2014explaining}, which are more efficient than traditional methods. 
Although the above perturbation-based attacks have achieved high success rates in attacking a target model, as shown in Figure \ref{fig:introduction}(b), they are severely limited by two challenging issues:
1) Existing works generally add or delete points learned by the gradient, or shift points along xyz directions by FGSM, which may result in messy distribution or outliers that deform the geometric structure. Therefore, the generated adversarial point clouds are often easily perceivable by humans and distinguished as attacks.
2) Most previous works generate adversarial examples in the white-box setting, where attackers have a perfect knowledge of the network structure and parameter weights. These attacks tend to overfit the target network and achieve high white-box success rates, but hardly remain malicious once they are transferred to attack a different victim model (\ie, in the black-box setting). 

To address the above two challenges\red{---large perturbation sizes and low transfer-based attack success rates}, we propose a \red{ {\it practical} and {\it robust} method for 3D point cloud attack, named} Imperceptible Transfer Attack (ITA).
\red{In particular, a practical and robust attack should satisfy two conditions: 1) A practical attack should deceive human/robot's eyes by keeping the geometric characteristics of the benign sample (imperceptibility); This also facilitates the 3D attack community to generate adversarial examples of high quality. 2) A robust attack should be less dependent on a specific victim network, and generalize better to different networks (transferability).
Moreover, the imperceptibility is connected to the transferability via the preservation of geometric characteristics. 
Specifically, to achieve the imperceptibility to humans, an adversarial point cloud should keep the geometric characteristics, \ie, the surface structure of the underlying manifold. 
That is, an imperceptible attack keeps the dependency among neighboring points so as to preserve the geometric contexts of benign point clouds, such as slowly-varying smooth surfaces and contours with large variations. 
Such crafted adversarial examples tend to be indistinguishable to different network models that generally learn the geometry of 3D shapes for classification, thus enhancing their transferability among these network models.
Hence, our ITA attack aims to generate qualified adversarial 3D point clouds with high imperceptibility and transferability.}

{Firstly}, to achieve the {\it imperceptibility} to humans, an adversarial point cloud should satisfy geometric characteristics such as piecewise smoothness \cite{hu2021overview,hu2014multiresolution,chao2015edge}, which exhibits slowly varying underlying surfaces separated by sharp edges.
Perturbing points along xyz directions would directly deform the geometric structure, thus deviating from such properties. 
\red{Therefore, we propose to constrain the perturbation along the normal direction in order to keep the geometric characteristics and the distribution of 3D shapes similar to the original benign one for improving the imperceptibility. The main reasons are twofolds:
1) In general, perturbing along the normal direction amounts to slightly zooming in or out of a point cloud, which is able to retain most geometric properties of the original 3D object, thus achieving the imperceptibility; 
2) Further, to keep the change in relative offsets between adjacent points small, we define a strictly bounded width to constrain the shifting magnitude of each point.
With this posed strict bound constraint, our perturbation method well keeps the position dependency among neighboring points and promotes the consistency of local curvatures between adversarial and benign point clouds, thus preserving the geometry of the original point cloud.}

{Secondly}, to improve the {\it transferability} of adversarial examples, most previous works in the image field \cite{xie2019improving,dong2019evading} train them to become robust against common image transformations under a fixed magnitude of distortion, which makes the adversarial examples difficult to resist against unknown distortions, thus leading to poor generalization. 
Hence, we develop a {\it learnable} transformation model that performs a wide spectrum of 3D transformations and infers the most harmful deformations to adversarial examples. 
Once the generated adversarial examples are able to resist such distortions introduced by the transformation model during adversarial learning, they are transferable to attack black-box models with higher success rates.

Specifically, given a point cloud, we pre-compute the normal vectors of all points and perturb each point along the direction of its normal within a strictly bounded width. During the generation of adversarial point clouds, we adversarially learn the transformation model to generate most harmful distortions to the adversarial examples while forcing the generated examples to be resistant to such transformations. In this way, our ITA attack is more imperceptible to humans with higher transferability.

\red{Further, since our ITA attack is robust to various existing defense methods, we propose an effective point cloud defense approach to motivate future research on defending against such imperceptible and transferable adversarial examples via a constrained adversarial training strategy.}
The goal is to adversarially train the black-box model from scratch without modifying the network structure, so that the trained model is able to recognize the clean and perturbed point clouds well in the inference stage. 
In particular, since the generated adversarial examples by the ITA attack preserve almost the same geometric information as the original one, previous works hardly distinguish these samples by directly measuring their point-wise distance and restoring the point clouds through pre-processing based defense in the data space, as illustrated in Figure \ref{fig:introduction}(c).
Therefore, we delve into their differences in the latent space where their feature representations tend to become more divergent in deeper embedding layers \cite{xie2019feature}, thus leading to different predictions of the model. 
To this end, we develop an adversarial-training-based defense method, in order to learn more discriminative point cloud representations that reduce the
divergence between the original and adversarial examples in the latent space. 
Inspired by the principle of enforcing the representations of intra-class point clouds to be closer and inter-class point clouds to be discriminative, we propose such a constraint loss in the latent space, which greatly enhances the robustness of black-box 3D models against unseen yet transferable attacks.

Our contributions mainly include:
\begin{itemize}
    \item We propose a novel Imperceptible Transfer Attack (ITA) for 3D point clouds, which generates qualified adversarial examples with high imperceptibility and transferability.
    \item The proposed ITA attack enhances the imperceptibility of adversarial point clouds by shifting each point in the direction of its normal vector within a strictly bounded width so as to keep geometric properties of the original point clouds.
    \item To enhance the transferability of adversarial point clouds, our ITA attack introduces a learnable transformation model and incorporates it into the white-box target model to generate adversarial examples in an adversarial learning manner. This enables the adversarial examples to resist against unknown distortions when sent to a black-box model.
    \item We further propose point cloud defense via an adversarial training strategy to improve the robustness of black-box models by introducing distance constraints in the latent space, which leads to discriminative feature representation learning and thus robust 3D models.  
    \item 
    Extensive evaluations demonstrate that our ITA attack achieves much higher success rates when transferred to unknown victim models than existing works, and is more imperceptible to humans. We also verify the effectiveness of the proposed adversarial-training-based defense method.
\end{itemize}
\section{Related Works}
\subsection{3D Point Cloud Classification}
Deep point cloud learning \cite{qi2017pointnet,qi2017pointnet++,wang2019dynamic,yang2018foldingnet,liu2019relation,rao2020global,te2018rgcnn,gao2020graphter} has emerged in recent years, which has diverse applications in many fields, such as object classification \cite{su2015multi,lei2020spherical}, scene segmentation \cite{graham20183d,hu2021learning}, and detection in autonomous driving \cite{chen2017multi,zhu2021cylindrical}. 
Among them, object classification \cite{su2015multi,yu2018multi,li2018pointcnn,zaheer2017deep,qi2017pointnet} is the most fundamental and important task, which learns representative information including both local details and global context of point clouds.
Early works attempt to classify point clouds by adapting deep learning models on the 2D grid \cite{su2015multi,yu2018multi}, which max-pool multi-view features into a global representation but suffer from high computation costs due to the usage of many 2D convolution layers. 
In order to directly learn the 3D structure and address the unorderness problem of point clouds, some works such as \cite{li2018pointcnn} transform the input points into a latent and potentially canonical order through an x-conv transformation, and then apply the typical convolutional operator on the transformed features.
Different from the above methods,
DeepSets \cite{zaheer2017deep} and PointNet \cite{qi2017pointnet} are the first to achieve end-to-end learning on point cloud classification and formulate a general specification for point cloud learning.
Specifically, PointNet learns point-wise features independently with several MLP layers and extracts the representative global features with a max-pooling layer.
Built upon PointNet, PointNet++ \cite{qi2017pointnet++} further captures fine local structural information from the neighborhood of each point by exploiting $k$ nearest neighbors.
DGCNN \cite{wang2019dynamic} also captures point-wise local geometric structure by developing an edge convolution operation with the graph dynamically constructed.
\red{PointTrans. \cite{zhao2021point} and  PointMLP \cite{ma2022rethinking} are recently developed based on the transformer and MLP, respectively.}
In this work, we focus on attacking popular point cloud classification models including PointNet, PointNet++, DGCNN, \red{PointTrans. and PointMLP}, and investigate the transferability of the generated adversarial examples among them.

\subsection{Existing Attacks \& Defenses on 2D Image Fields}
Deep neural networks are vulnerable to adversarial examples, which has been extensively explored in the 2D image field \cite{moosavi2016deepfool,moosavi2017universal}.
According to the type of victim models, there are typically two families of attacks in the image domain: 1) white-box attack: the attackers have a perfect knowledge of the target model and directly generate adversarial examples against it; 2) black-box attack: the attackers attack a remote victim model whose network structure and parameters are unknown.
For the white-box attack, previous works often follow the point-wise perturbation scheme \cite{goodfellow2014explaining,kurakin2016adversarial,madry2017towards,carlini2017towards,carlini2017towards,szegedy2013intriguing}, which searches for pixel-level perturbations on clean images by optimizing the misclassification loss and calculating corresponding gradients. 
The Fast Gradient Sign Method (FGSM) \cite{goodfellow2014explaining} is the first to propose a single-step gradient update strategy to attack the network by moving the benign images towards a wrong decision boundary. The C\&W attack \cite{carlini2017towards} further proposed to include both the misclassification loss and a loss constraining the magnitudes of adversarial noise into the optimization objective.
Regarding the black-box attack, query-based methods \cite{bhagoji2018practical,guo2019simple,li2020regional} generally utilize excessive instances to query the victim model for collecting their feedback information and generating corresponding adversarial examples. These works require high query costs and make the attacks more detectable. 
Transfer-based methods \cite{dong2018boosting,dong2019evading,lin2020nesterov} first generate adversarial images against a target model, and then transfer the adversarial samples to attack a remote victim model. 
In this paper, we mainly focus on the transfer-based attack under the black-box setting, and generate adversarial examples on more challenging 3D models.

In order to mitigate the threat of potential attacks, there are a variety of defense approaches proposed to detect adversarial examples on the image domain. 
The widely used transformation-based methods \cite{guo2017countering,xie2017mitigating,liao2018defense,xu2017feature,jia2019comdefend,cohen2019certified} typically adopt various kinds of transformations (\eg, resizing and padding, cropping, compression) to eliminate adversarial perturbations before sending them to the victim model. 
Besides, the adversarial training strategy \cite{goodfellow2014explaining,kurakin2016adversarial,tramer2017ensemble} is the most promising defense that improves the robustness of the target model by training from scratch with both clean and adversarial data. 

\subsection{Existing Attacks \& Defenses on 3D Point Clouds}
Following previous studies on the 2D image field, many works \cite{xiang2019generating,wicker2019robustness,zhang2019adversarial,zheng2019pointcloud,tsai2020robust,zhao2020isometry,zhou2020lg} adapt adversarial attacks into the 3D vision community. 
Xiang \et \cite{xiang2019generating} proposed point generation attacks by adding a limited number of synthesized points/clusters/objects to a point cloud, and show its effectiveness on attacking the PointNet model \cite{qi2017pointnet}. 
Zhang \et \cite{zhang2019adversarial} utilize gradient-guided attack methods to explore point modification, addition, and deletion attacks.
Their goal is to add or delete key points, which can be identified by calculating the label-dependent importance score referring to the calculated gradient.
Recently, more works \cite{liu2019extending,zhang2019defense,tsai2020robust} adopt point-wise perturbation by changing their xyz coordinates, which are more effective and efficient. 
Liu \et \cite{liu2019extending} modify the FGSM strategy to iteratively search the desired pixel-wise perturbation.
Tsai \et \cite{tsai2020robust} adapt the C\&W strategy to generate adversarial examples on point clouds and proposed a perturbation-constrained regularization in the overall loss function. They also deform the mesh-level offsets by modifying the gradient direction.
Besides, some works \cite{lee2020shapeadv,tian2021poisoning} attack point clouds in the feature space or utilize the backdoor attack scheme.
However, the generated adversarial point clouds of all above methods often result in messy distribution or outliers, which is easily perceivable by humans. 
Although Wen \et \cite{wen2020geometry} improve the imperceptibility of adversarial attacks by enforcing a geometry-aware constraint, the generated adversarial samples sometimes also deform local surfaces and are still noticeable to humans. 
In this paper, we develop a directional perturbation strategy to shift each point along its normal vector within a strictly bounded width for enhancing the imperceptibility. 
Moreover, we delve into the transferability of adversarial examples by firstly learning the most harmful deformation and then enforcing the adversarial examples to resist it.

Existing defense methods on point clouds \cite{liu2019extending,zhou2019dup,zhang2019adversarial,dong2020self,wu2020if,liu2021pointguard} proposed to defend against adversarial attacks.
Simple random sampling (SRS) \cite{zhang2019adversarial} is a widely used strategy to firstly downsample the point cloud into a smaller set, and then take the predicted class of such set as the result of the original point cloud.
Zhou \et \cite{zhou2019dup} proposed both statistical outlier removal (SOR) and Denoiser and UPsampler Network (DUP-Net) defenses, which target at detecting and discarding outliers before feeding them into the 3D models.
Dong \et \cite{dong2020self} introduced the gather-vector to distinguish the clean and adversarial local features, and then filter out the latter ones in the final global feature.
Wu \et \cite{wu2020if} proposed the Implicit Function Defense (IF-Defense) to optimize and reconstruct the adversarial point cloud with both geometry-aware and distribution-aware losses.
Liu \et \cite{liu2021pointguard} utilized certified defense to build the first robust 3D classification model that is able to effectively defend against most adversarial attacks.
In this work, we evaluate the adversarial robustness of our ITA attack by employing the above defense methods. 
We also propose a defense method against imperceptible and transferable adversarial examples via a robust adversarial training strategy.
\section{Imperceptible Transfer Attack}
In this section, we elaborate on the proposed Imperceptible Transfer Attack (ITA) for point clouds. 
We first introduce the general formulation of generating adversarial point clouds, and then provide detailed descriptions of enhancing both the imperceptibility and transferability of the adversarial point clouds, respectively. In the end, we present our final optimization formulation to generate adversarial examples.

\subsection{Problem Statement}
A point cloud consists of an unordered set of points $\bm{P}=\{\bm{p}_i\}_{i=1}^n \in \mathbb{R}^{n \times 3}$ sampled from the surface of a 3D object or scene, where each point $\bm{p}_i \in \mathbb{R}^{3}$ is a vector that contains the coordinates $(x,y,z)$ of point $i$, and $n$ is the number of points. 

In the point cloud classification task that we focus on, given a point cloud $\bm{P}$ as input, a learned classifier $f(\cdot)$ predicts a label $y=f(\bm{P}) \in {\sY}, {\sY}= \{1,2,3,...,C\}$ that represents the class of the original 3D object underlying the point cloud, where $C$ is the number of classes. 
To attack such classification model, the general objective \cite{zhang2019adversarial,xiang2019generating,wen2020geometry} is to find a perturbation $\bm{\Delta} \in \mathbb{R}^{n \times 3}$ that generates an adversarial example as $\bm{P}' = \bm{P} + \bm{\Delta}$. Similar to attacks in the image domain, there are generally two types of attacks: 1) the {\it untargeted} attack \cite{su2019one,goodfellow2014explaining} generates adversarial example $\bm{P}'$ such that $f(\bm{P}') \neq y$; 2) the {\it targeted} attack \cite{szegedy2013intriguing,papernot2016limitations} generates $\bm{P}'$ such that $f(\bm{P}')=y'$, where $y' \in \sY$ but $y' \neq y$. 

Since the {\it targeted} attack is more challenging than the {\it untargeted} one, we focus on the {\it targeted} attack in this paper. 
The objective is to learn an adversarial example $\bm{P}'$ such that 1) $\bm{P}'$ is misclassified as a targeted false class $y'$; and 2) $\bm{P}'$ is similar to the original point cloud $\bm{P}$.   
In particular, we formulate the targeted attack as the following optimization problem via searching an appropriate $\bm{\Delta}$:
\begin{equation}
\label{eq:1}
\begin{split}
    &\min_{\bm{\Delta}} \mathcal{L}_{mis}(f(\bm{P}'),y') + \lambda \mathcal{L}_{reg}(\bm{P}',\bm{P}), \\
    &\text{s.t.}~~ \ \bm{P}' = \bm{P} + \bm{\Delta},
\end{split}
\end{equation}
where $\mathcal{L}_{mis}$ is the cross-entropy loss to promote the misclassification of $\bm{P}'$ to $y'$, $\mathcal{L}_{reg}$ is a regularization term that minimizes the distance between $\bm{P}'$ and $\bm{P}$, and $\lambda$ is a parameter to strike a balance between the two terms in the objective. 

We endeavor to develop an imperceptible
attack by proposing a novel perturbation approach to preserve the underlying structure of the original point cloud. 
Further, we build an adversarial transformation model to escalate the transferability of adversarial examples for attacking a black-box model with high success rates. 
We discuss the proposed ITA attack to enhance the imperceptibility and transferability in order as follows.  

\begin{figure}
    \centering
    \includegraphics[width=0.48\textwidth]{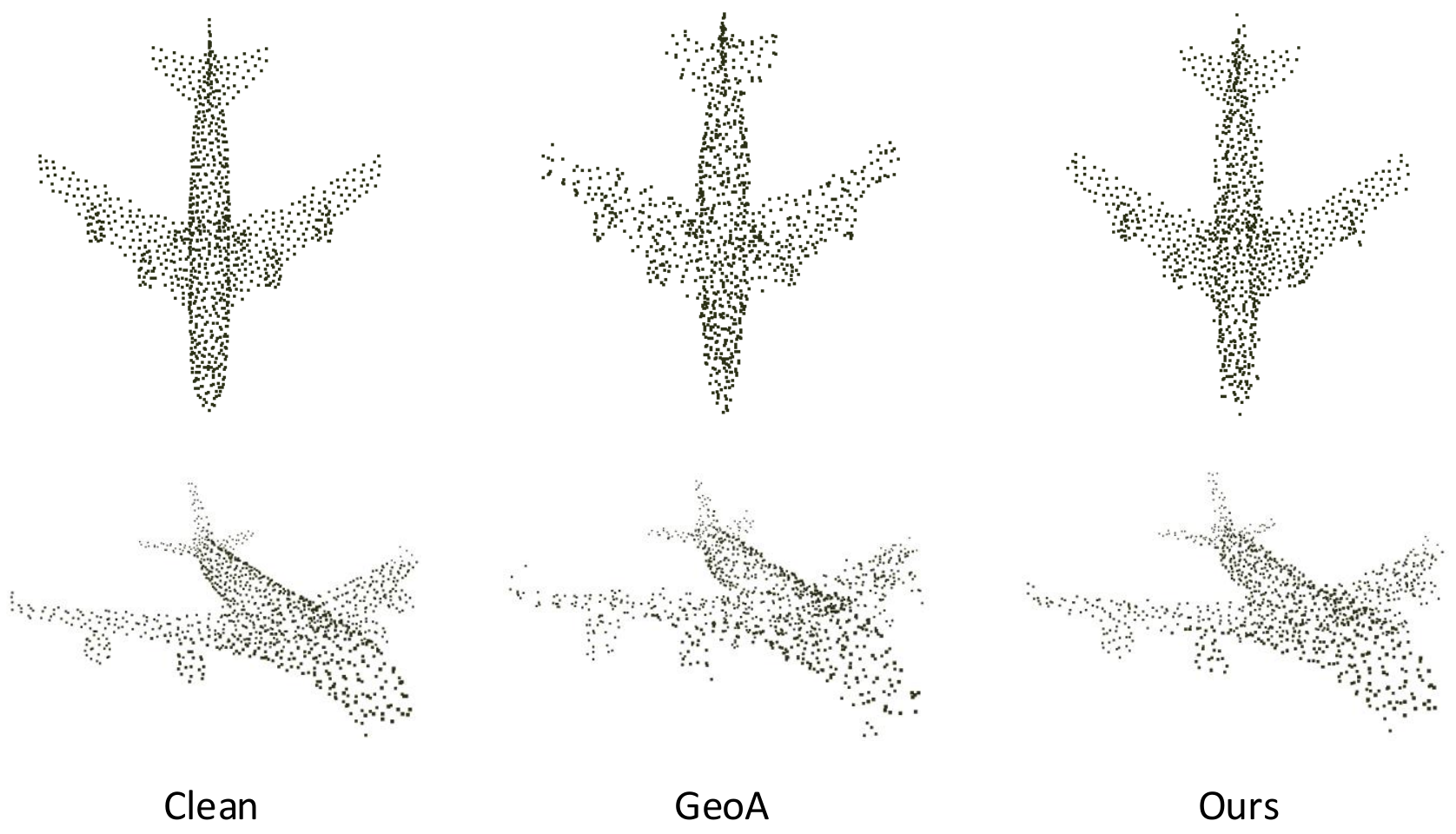}
    \caption{Visualization results of the generated adversarial examples of different methods. Compared to the previous most imperceptible GeoA attack, our adversarial examples are more imperceptible to humans without outlier or uneven local distribution.}
    \label{fig:bound1}
\end{figure}

\subsection{Enhancing the Imperceptibility}
While most existing attack methods perturb points away from the underlying surface that deforms the structure, we aim to improve the imperceptibility by keeping the geometric characteristics of point clouds. 
Although the geometry-aware objective attack (GeoA) \cite{wen2020geometry} attempts to retain the structural properties by promoting the consistency of local curvatures between the clean and adversarial point clouds, GeoA still results in uneven distributions of local surfaces due to the point-wise perturbation along xyz dimensions, which is also perceptible to humans as shown in Figure \ref{fig:bound1}. 
To this end, we propose a directional perturbation method to enforce point-wise shifting along the normal direction in order to alleviate the change in local distributions, and further constrain the learned perturbation within a certain bound to preserve the geometric characteristics.

\subsubsection{Directional Perturbation}
To ensure the geometric consistency between the perturbed point cloud $\bm{P}'$ and the original point cloud $\bm{P}$, we aim to preserve the geometric characteristics of the underlying local surfaces around each point in $\bm{P}$.
Therefore, we consider to perturb each point along the direction of its normal vector. 

Specifically, to acquire the normal vector for each point $\bm{p}_i \in \bm{P}$ with its $k$-nearest neighbors $\mathcal{N}_{\bm{p}_i}$, we first follow a traditional strategy \cite{hoppe1992surface} to compute the symmetric and positive semi-definite matrix $\bm{S}_{\bm{p}_i} \in \mathbb{R}^{3 \times 3}$ as:
\begin{equation}
\label{eq:2}
    \bm{S}_{\bm{p}_i} = \sum_{\bm{p}_j \in \mathcal{N}_{\bm{p}_i}, j \neq i} (\bm{p}_j - \bm{p}_i) \otimes (\bm{p}_j - \bm{p}_i),
\end{equation}
where $\otimes$ denotes the operator of outer product.
Then, we perform the eigen-decomposition of $\bm{S}_{\bm{p}_i}$ to obtain three orthonormal eigenvectors $\bm{e}^1_{\bm{p}_i},\bm{e}^2_{\bm{p}_i},\bm{e}^3_{\bm{p}_i}$, 
where we choose the unit normal vector $\bm{u}_{\bm{p}_i}$ to be either $\bm{e}^3_{\bm{p}_i}$ or $-\bm{e}^3_{\bm{p}_i}$, while $\bm{e}^1_{\bm{p}_i},\bm{e}^2_{\bm{p}_i}$ define the tangent plane.

\begin{figure}
    \centering
    \includegraphics[width=0.46\textwidth]{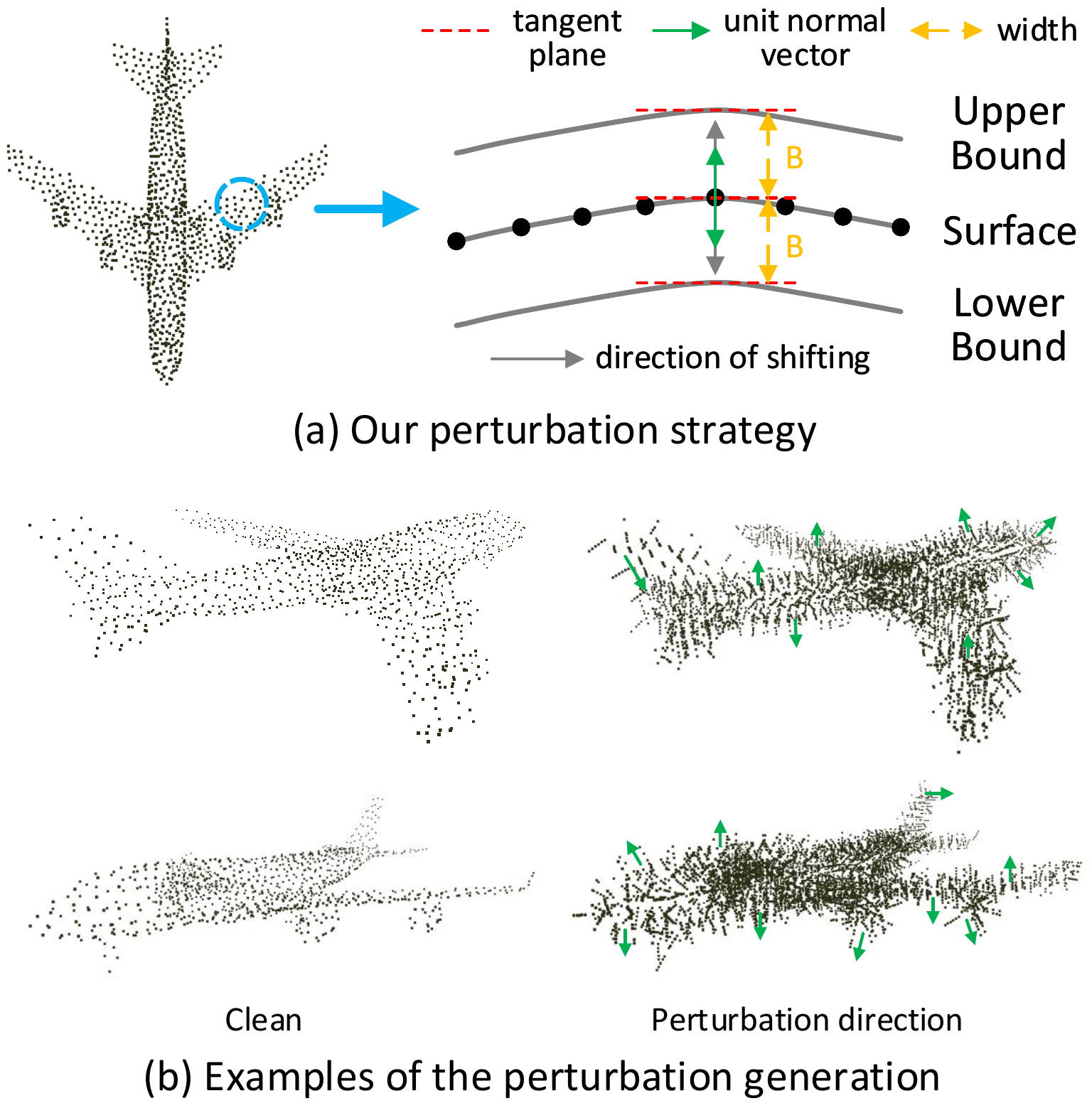}
    \caption{(a) Illustration of our perturbation strategy, where we only perturb each point along the direction of its normal vector within a strictly bounded width. (b) Visualization examples of the perturbation generation, where each point is perturbed along its normal vector.}
    \label{fig:bound2}
\end{figure}

After obtaining the unit normal vector of each point, as shown in Figure \ref{fig:bound2}, we propose to perturb each point along the direction of its normal vector to generate the adversarial examples. The optimization function in Eq.~(\ref{eq:1}) is reformulated as follows:
\begin{equation}
\begin{split}
    &\min_{\bm{\Delta}} \mathcal{L}_{mis}(f(\bm{P}'),y') + \lambda \mathcal{L}_{reg}(\bm{P}',\bm{P}), \\
    &\text{s.t.}~~ \ \bm{P}' = \bm{P} + \bm{\Delta} \cdot \bm{U},
\end{split}
\end{equation}
where perturbation $\bm{\Delta}$ is re-defined as the shifting length along the normal direction with dimension $\mathbb{R}^{n \times 1}$ for all the $n$ points, and $\bm{U}=\{\bm{u}_{\bm{p}_i}\}_{i=1}^n \in \mathbb{R}^{n \times 3}$. 
Since a large length $\bm{\Delta}$ would damage the geometric properties and lead to outliers, we further constrain the shifting length within a strictly bounded width $B \in [0,1]$ as follows:
\begin{equation}
\begin{split}
    &\min_{\bm{\Delta}} \mathcal{L}_{mis}(f(\bm{P}'),y') + \lambda \mathcal{L}_{reg}(\bm{P}',\bm{P}), \\
    & \text{s.t.}~~ \bm{P}' = \bm{P} + \bm{\Delta} \cdot \bm{U}, ~~\mid {\Delta}_i \mid < B, i=1,...,n. 
\end{split}
\end{equation}
The tighter the width $B$ is, the more geometric characteristics will be retained.
The optimal width value is empirically assigned according to the ablation study in Section~\ref{exp:bound}.
In this way, the generated adversarial point clouds are able to keep the geometric characteristics of the original one, thus alleviating the issue of uneven local distribution and outliers.

\subsubsection{Regularization Loss}
We deploy two distance metrics of point clouds as the regularization loss $\mathcal{L}_{reg}(\bm{P}',\bm{P})$---Hausdorff distance \cite{huttenlocher1993comparing} and Chamfer distance \cite{fan2017point}. 

In particular, we first employ the Hausdorff distance to measure the {\it maximum} squared Euclidean distance among all nearest point pairs between the clean point cloud $\bm{P}$ and its adversarial example $\bm{P}'$.
Specifically, given these two point clouds $\bm{P}'$ and $\bm{P}$, the Hausdorff distance is defined as:
\begin{equation}
    \mathcal{L}_h(\bm{P}',\bm{P}) = \max_{\bm{p}'_i \in \bm{P}'} \min_{\bm{p}_j \in \bm{P}} \parallel \bm{p}'_i - \bm{p}_j \parallel_2^2.
\end{equation}
This distance metric finds the nearest point from the original point cloud for each adversarial point and computes the maximum squared Euclidean distance among all such nearest point pairs, which is sensitive to outlier points.

Also, we utilize the Chamfer distance to compute the {\it average} distance between point clouds $\bm{P}'$ and $\bm{P}$ as:
\begin{equation}
    \mathcal{L}_c(\bm{P}',\bm{P}) = \frac{1}{n} \sum_{\bm{p}'_i \in \bm{P}'} \min_{\bm{p}_j \in \bm{P}} \parallel \bm{p}'_i - \bm{p}_j \parallel_2^2.
\end{equation}

Hence, the objective to generate imperceptible adversarial point cloud $\bm{P}'$ is formulated as:
\begin{equation}
\label{eq:7}
\begin{split}
    &\min_{\bm{\Delta}} \mathcal{L}_{mis}(f(\bm{P}'),y') + \lambda_1 \mathcal{L}_{h}(\bm{P}',\bm{P}) + \lambda_2 \mathcal{L}_{c}(\bm{P}',\bm{P}), \\
    &\text{s.t.}~~ \bm{P}' = \bm{P} + \bm{\Delta} \cdot \bm{U}, ~~\mid {\Delta}_i \mid < B, i=1,...,n,
\end{split}
\end{equation}
where $\lambda_1, \lambda_2$ are two parameters for the balance among the terms in the objective.

\begin{figure*}[t!]
    \centering
    \includegraphics[width=1.0\textwidth]{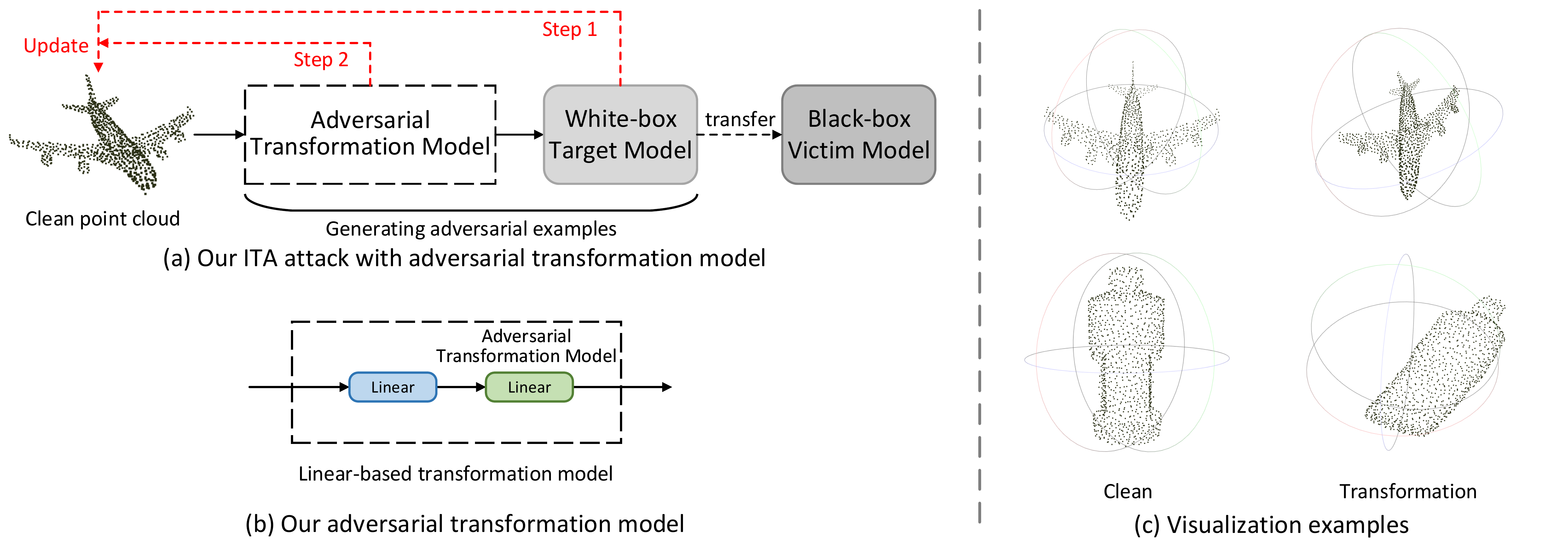}
    \caption{(a) Illustration of our ITA attack with an adversarial transformation model, where we adopt a two-step adversarial learning strategy to train the transformation model. (b) \red{To implement our adversarial transformation model, we directly deploy two linear layers to imitate natural point-wise transformations.} (c) Examples of the output point clouds from the learned \red{adversarial} transformation models.}
    \label{fig:transformation}
\end{figure*}

\subsubsection{\red{Comparison with the Curvature Loss}}
\red{Although the GeoA \cite{wen2020geometry} attempts to keep the geometry information via their proposed curvature loss function, they shift and optimize each point {\it along xyz directions} to search both the desirable geometry-aware perturbation direction and corresponding perturbation size.
Instead, our attack starts from {\it along which direction to add the geometry-aware perturbation} and directly constrains the perturbation direction along the normal vector so as to preserve the geometry information (keeping the position dependency among neighboring points and thus promoting the consistency of local curvatures). 
This is more effective to generate more imperceptible adversarial examples, and thus leads to a smaller Curvature Loss compared to the original GeoA paper even though we do not explicitly optimize for a Curvature Loss during the attack.}

\subsection{Enhancing the Transferability}

To further enhance the transferability of the generated adversarial examples, we aim to enforce them to be robust against various transformations (\eg, affine transformations including translation, rotation and shearing), which generally induce deformations in the underlying structure. 
The insight is that, if the adversarial examples can resist the toughest deformations, they are also able to survive under weaker distortions when transferred to an unknown model. 

To this end, as shown in Figure \ref{fig:transformation} (a) and (b), we develop a deep-learning based transformation model $T(\cdot)$ to learn the most destructive transformations to each adversarial point cloud. 
Particularly, we propose to implement model $T(\cdot)$
\red{by a combination of multiple linear layers, which learn to generate a composition of multiple simple transformations for point clouds.}
During the training, we integrate the transformation model with a fixed white-box target model to be a new victim model, and utilize a two-step adversarial learning strategy to iteratively attack such model while defending the attack by updating the transformation model. More details can be found in Figure \ref{fig:transformation}(a).

Specifically, to learn a robust transformation model, we deploy a recent adversarial learning scheme \cite{goodfellow2020generative} to generate destructive distortions for adversarial examples. 
In the first step, we aim to \textit{search the adversarial point cloud} $\bm{P}'$ by attacking a single classification model $f(\cdot)$ as well as the model $f(T(\cdot))$ combined from both transformation and classification models as:
\begin{equation}
\label{eq:8}
\begin{split}
    &\min_{\bm{\Delta}} \mathcal{L}_{mis}(f(\bm{P}'),y') + \alpha \mathcal{L}_{mis}(f(T(\bm{P}')),y') \\
    &\qquad + \lambda_1 \mathcal{L}_{h}(\bm{P}',\bm{P}) + \lambda_2 \mathcal{L}_{c}(\bm{P}',\bm{P}), \\
    &\text{s.t.}~~ \bm{P}' = \bm{P} + \bm{\Delta} \cdot \bm{U}, ~~\mid {\Delta}_i \mid < B, i=1,...,n,
\end{split}
\end{equation}
where the second item endeavors to make the adversarial example remain malicious under the transformation network, and $\alpha$ is a weighting parameter. 

\begin{algorithm}[t!] 
\caption{Adversarial learning the transformation model} 
\label{alg:adv} 
{\bf Input:} The clean point cloud $\bm{P}$, the trained classification model $f(\cdot)$, point-wise unit normal vectors $\bm{U}$, the iteration numbers $L_1,L_2,L_3$.
\begin{algorithmic}[1]
\STATE Randomly initialize the perturbation $\bm{\Delta}$ and model $T(\cdot)$
\STATE Initialize the perturbed point cloud $\bm{P}' = \bm{P} + \bm{\Delta} \cdot \bm{U}$
\STATE {\bf for} iteration $l_1 \leftarrow 1$ to $L_1$ {\bf do}
\STATE \qquad {\bf for} iteration $l_2 \leftarrow 1$ to $L_2$ {\bf do} \textit{(step 1)}
\STATE \qquad \qquad Learn $\bm{\Delta}$ by Eq.~(\ref{eq:8})
\STATE \qquad \qquad Update $\bm{P}' = \bm{P} + \bm{\Delta} \cdot \bm{U}$
\STATE \qquad {\bf end for}
\STATE \qquad {\bf for} iteration $l_3 \leftarrow 1$ to $L_3$ {\bf do} \textit{(step 2)}
\STATE \qquad \qquad Learn $T(\cdot)$ by Eq.~(\ref{eq:9})
\STATE \qquad \qquad Update the parameters of $T(\cdot)$
\STATE \qquad {\bf end for}
\STATE {\bf end for}
\STATE {\bf Return} the trained transformation model $T(\cdot)$
\end{algorithmic}
\end{algorithm}

\begin{figure*}[t!]
    \centering
    \includegraphics[width=1.0\textwidth]{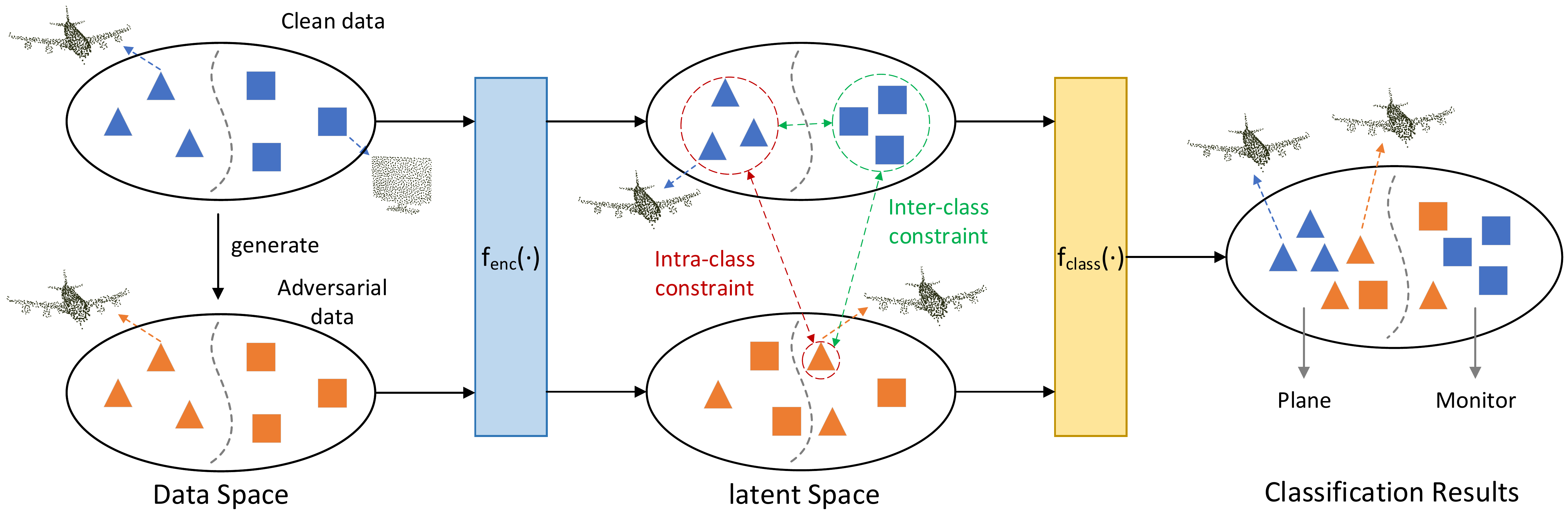}
    \caption{Illustration of our proposed intra- and inter-class constraints in the adversarial training process for point cloud defense. Given the clean point cloud data, we first generate corresponding adversarial examples, and then feed both of them into the 3D model for discriminative representation learning. The intra-class constraint is proposed to reduce the divergence between clean and adversarial point clouds of the same object. Meanwhile, the inter-class constraint aims to enlarge the difference among point clouds belonging to different objects.}
    \label{fig:defense}
\end{figure*}

In the second step of adversarial learning, we \textit{train and update the transformation model $T(\cdot)$} to defend such adversarial example $\bm{P}'$ by making $\bm{P}'$ less destructive to the target classifier, while maintaining the characteristics of the original $\bm{P}$ by accurately predicting its object class. 
We formulate this loss function as follows:
\begin{equation}
\label{eq:9}
\begin{split}
    &\min_{T(\cdot)} \mathcal{L}_{mis}(f(T(\bm{P}')),y) + \beta \mathcal{L}_{mis}(f(T(\bm{P})),y) \\
    &\qquad + \lambda_3 \mathcal{L}_{c}(\bm{P}',T(\bm{P}')),
\end{split}
\end{equation}
where the third item employs the Chamfer distance to constrain the distortion, and $\beta,\lambda_3$ are weighting parameters.
We alternate the searching for the adversarial example and the training of the adversarial transformation network in an iterative process, as presented in Algorithm \ref{alg:adv}.

We provide several visualization examples output from the learned \red{adversarial} transformation model in Figure \ref{fig:transformation}(c). 
The first column shows the clean point clouds, while the second column illustrates the corresponding point clouds transformed by the \red{Linear-based transformations}. 
\red{We observe that our Linear-based adversarial transformation model converges at a combination of multiple simple transformations (\ie, translation, rotation, shearing).}
This demonstrates that our adversarial transformation models possess the capacity to generate distortions of point clouds for the enhancement of transferability.

To evaluate the transferability of our generated adversarial point clouds, we directly feed them into remote black-box victim models to compute the attack success rates, which will be demonstrated in Section~\ref{exp:transfer}.

\subsection{Final Problem Formulation of the ITA}
After acquiring the trained transformation model $T(\cdot)$, we integrate it with the trained classification model $f(\cdot)$, leading to a new white-box target model $f(T(\cdot))$ to attack. 
Hence, we finally formulate the proposed ITA model as:
\begin{equation}
\label{eq:10}
\begin{split}
    &\min_{\bm{\Delta}} \mathcal{L}_{mis}(f(T(\bm{P}')),y') + \lambda_1 \mathcal{L}_{h}(\bm{P}',\bm{P}) + \lambda_2 \mathcal{L}_{c}(\bm{P}',\bm{P}), \\
    &\text{s.t.}~~ \bm{P}' = \bm{P} + \bm{\Delta} \cdot \bm{U}, ~~\mid {\Delta}_i \mid < B, i=1,...,n,
\end{split}
\end{equation}
which promotes both the imperceptibility and transferability of generated adversarial point clouds. 

\section{Defense against imperceptible \& transferable adversarial point clouds}

So far, we have generated both imperceptible and transferable adversarial point clouds by our ITA attack. Existing defenses on point clouds \cite{liu2019extending,zhou2019dup,zhang2019adversarial,dong2020self,wu2020if,liu2021pointguard} generally employ empirical operations (\eg, point cloud sampling, point removal, denoising) to reconstruct the geometry of the point cloud, which severely rely on time-consuming point cloud pre-processing and are ineffective to the proposed ITA since we preserve the underlying geometric structure during the attack process\footnote{Experiments in Section~\ref{exp:defense} also validate this intuition.}. Moreover, some of these defense methods \cite{dong2020self,liu2021pointguard} directly modify the network structure of the target model to defend attacks, which are limited to the white-box setting and fail to guide the black-box victim model to recognize transfer-based adversarial examples. 

To address the aforementioned limitations, we develop a generic defense approach to improve the robustness of any 3D model (white-box or black-box) by adopting adversarial training \cite{goodfellow2014explaining} and introducing a novel constraint loss in latent space. In particular, our defense method is plug-and-play for any 3D model without modifying network structure.

Given the clean point cloud data, our ITA attack generates corresponding adversarial point clouds that preserve most geometric information and are imperceptible. To make our defense adaptable, we deploy the adversarial training strategy to alternatively train the clean and adversarial data for improving the robustness and generalization of the 3D model. In particular, as shown in Figure \ref{fig:defense}, we propose to capture and constrain the similarity of intra- and inter-class benign/crafted point clouds in the latent space with a novel constraint loss.
The idea is that, although both clean and corresponding adversarial point clouds are visually similar and only differ in trivial perturbation, their feature representations diverge in the latent space \cite{xie2019feature}. This is the key reason why a 3D model would misclassify these adversarial examples. If the 3D model is able to learn similar representations for the clean and adversarial point clouds of the same instance, it will be more robust to various attacks. Hence, we are motivated to develop both intra- and inter-class constraint losses for robust feature representation learning of each point cloud.

\subsection{Intra-class Constraint}
We propose an intra-class constraint loss to reduce the divergence between the clean point cloud and corresponding adversarial samples (or other clean point cloud instances) of the same object. 
To access their feature representations in the latent space, we divide the 3D model $f(\cdot)$ into two parts: the feature encoder module $f_{enc}(\cdot)$ and the classification layer $f_{class}(\cdot)$, where $f(\cdot)=f_{class}(f_{enc}(\cdot))$. 
For example, for the PointNet model \cite{qi2017pointnet}, we take its global feature as the cloud-wise representation. The layers ahead of the max-pooling layer are taken as the feature encoder module, and the subsequent linear layers are taken as the classification module.

Given a clean point cloud $\bm{P}$ with label $y$, we first generate its corresponding adversarial sample $\bm{P}'$ by any adversarial attack (\eg, a simple FGSM strategy, the ITA, \etc), and sample the other $J$ numbers of point clouds $\{\bm{P}^b_{j,c=y}\}_{j=1}^{j=J}$ belonging to the same class $y$ with $\bm{P}$ in each minibatch $b$.
Then, we minimize the distance in their cloud-wise feature representations in the latent space as:
\begin{equation}
\label{eq:11}
\begin{split}
    \mathcal{L}_{intra} = \ & \omega_1 || f_{enc}(\bm{P}) - f_{enc}(\bm{P}') ||_2 \\
    & + \frac{\omega_2}{J} \sum_{j=1}^J || f_{enc}(\bm{P}) - f_{enc}(\bm{P}^b_{j,c=y}) ||_2,
\end{split}
\end{equation}
where $\omega_1,\omega_2$ are weighting parameters.

\subsection{Inter-class Constraint}
Further, we propose an inter-class constraint loss to enlarge the divergence between two clean point clouds of different classes or between the clean and adversarial point clouds of different classes. 
Given a clean point cloud $\bm{P}$, we first sample other point clouds $\{\bm{P}^b_{j,c}\}_{j=1,c=1,c \neq y}^{j=J,c=C,c \neq y}$ belonging to a different class with $\bm{P}$ in each minibatch, and generate targeted adversarial examples $\{\bm{P}'_{c}\}_{c=1,c\neq y}^{c=C,c\neq y}$ on $\bm{P}$. To differentiate the feature representations of inter-class point clouds, we formulate the constraint loss as:
\begin{equation}
\label{eq:12}
\begin{split}
    \mathcal{L}_{inter} = \ & -\frac{\omega_3}{(C-1) \cdot J} \sum_{c=1,c\neq y}^C \sum_{j=1}^J || f_{enc}(\bm{P}) - f_{enc}(\bm{P}^b_{j,c}) ||_2 \\
    & -\frac{\omega_4}{(C-1) \cdot J} \sum_{c=1,c\neq y}^C \sum_{j=1}^J || f_{enc}(\bm{P}'_c) - f_{enc}(\bm{P}^b_{j,c}) ||_2,
\end{split}
\end{equation}
where $\omega_3,\omega_4$ are weighting parameters.

\subsection{The Overall Loss Function}
\begin{figure}[t!]
    \centering
    \includegraphics[width=0.46\textwidth]{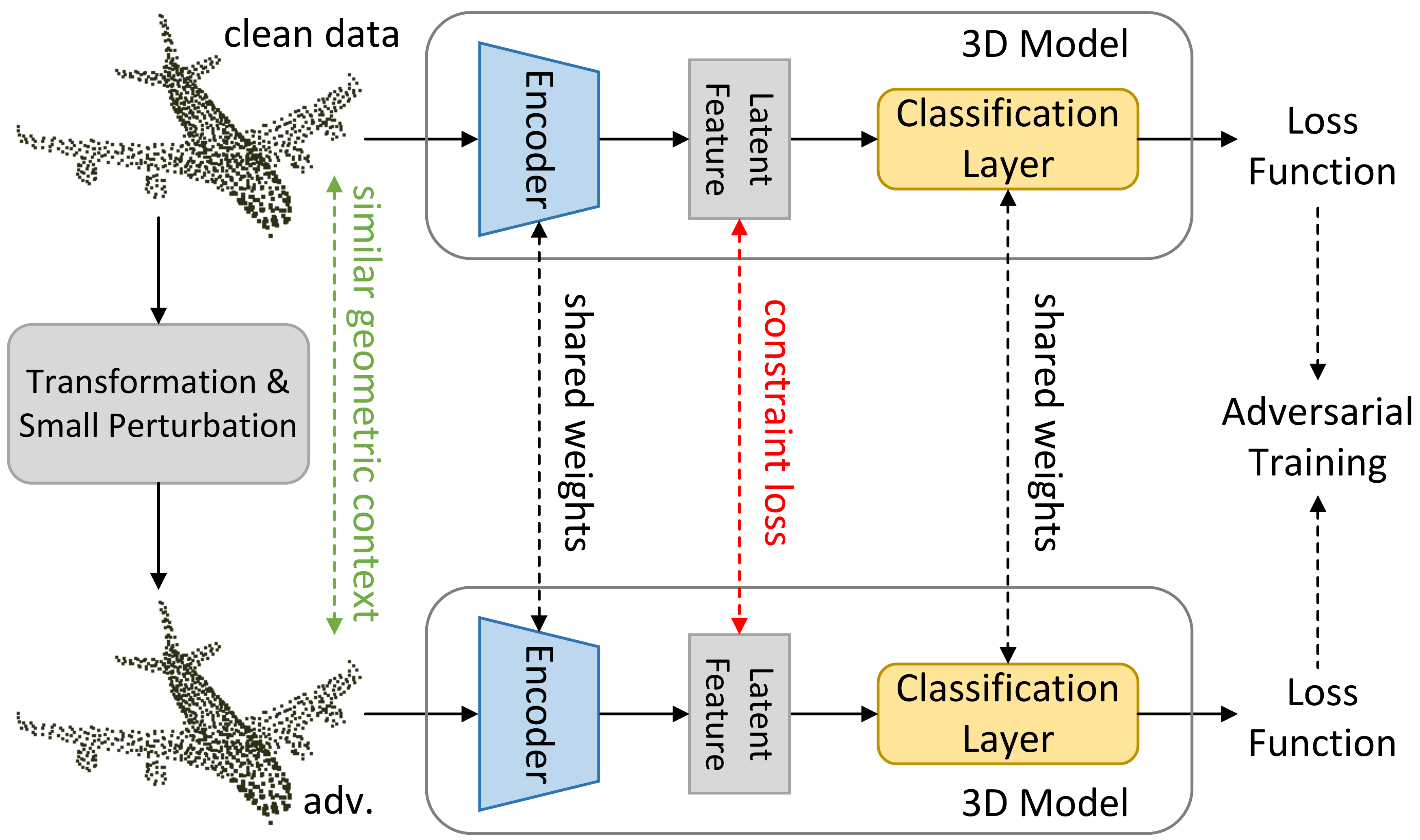}
    \caption{The framework of the proposed adversarial-training-based defense with similarity constraints in the latent space to improve the robustness of black-box models.}
    \label{fig:adversarial}
\end{figure}

Leveraging both intra- and inter-class constraint losses, the 3D model is able to learn robust feature representations where the imperceptible adversarial point cloud has smaller distance to its ground-truth point cloud than to the target-attack one.
Therefore, we formulate the overall loss function for the adversarial training process as follows:
\begin{equation}
\label{eq:13}
    \mathcal{L}_{overall} = \mathcal{L}_{class} + \omega_5 \mathcal{L}_{intra} + \omega_6 \mathcal{L}_{inter},
\end{equation}
where $\mathcal{L}_{class}$ employs the cross-entropy loss to train the classification layer $f_{class}(\cdot)$ with $\bm{P},\bm{P}'$ and their ground-truth class labels.
$\omega_5,\omega_6$ are weighting parameters to strike a balance among the three losses. 
The overall framework of the proposed point cloud defense is illustrated in Figure \ref{fig:adversarial}. 

\subsection{\red{Comparison with Relevant Learning Models}}
\red{Interestingly, both the triplet loss \cite{schroff2015facenet} and our proposed defense loss aim to learn discriminative representations by the principle of enforcing the features of similar elements to be proximal and dissimilar elements to be different. 
However, the triplet loss works on an {\it instance} level, which enforces the representation of an image to be proximal with those of its transformations and to be distant with those of any other images. 
In contrast, our method works on a {\it class} level, which enforces the intermediate representations of each class (containing both benign and adversarial data) to be close and well separated with those in other classes.} 

\red{Although there are two related works \cite{mustafa2019adversarial,wan2018rethinking} in image fields that also study in the context of adversarial robustness, \cite{mustafa2019adversarial} forces the features for each \textit{class} to lie inside a convex polytope that is maximally separated from the polytopes of other classes. 
Different from it, we directly pull the benign and adversarial samples of the intra-class closer and push the inter-class further.
Besides, \cite{mustafa2019adversarial} simply encompasses the untargeted adversarial sample space within a perturbation budget \textit{without} adversarial example generation. Instead, we generate and utilize more challenging targeted attack samples in our learning process.
Comparing our strategy with \cite{wan2018rethinking}, \cite{wan2018rethinking} develops a novel loss function for improving the robustness of the model without using adversarial training. Different from it, we adopt adversarial training strategy to discriminate the representations of both benign and adversarial samples.
Furthermore, both \cite{mustafa2019adversarial} and \cite{wan2018rethinking} are evaluated to work well on image datasets. Instead, we make the attempt to introduce appropriate intra- and inter-class constraints in 3D data for defending the 3D attacks in the feature space.}
\section{Experiments}
\subsection{Dataset and 3D Models}
\subsubsection{Dataset}
We adopt the point cloud benchmark ModelNet40 \cite{wu20153d} dataset in all the experiments. 
This dataset contains 12,311 CAD models from 40 most common object categories in the world. 
Among them, 9,843 objects are used for training and the other 2,468 for testing. 
Following previous works \cite{qi2017pointnet}, we uniformly sample $n=1,024$ points from the surface of each object, and re-scale them into a unit ball. 
For adversarial point cloud attacks, we follow \cite{xiang2019generating} and randomly select 25 instances for each of 10 object categories in the ModelNet40 testing set, which can be well classified by the classifiers of interest.

\subsubsection{3D Models}
Following previous works, we select \red{five} commonly used point cloud classification networks in 3D computer vision community as the victim models, \ie, PointNet \cite{qi2017pointnet}, PointNet++ \cite{qi2017pointnet++}, DGCNN \cite{wang2019dynamic}, \red{PointTrans. \cite{zhao2021point} and  PointMLP \cite{ma2022rethinking}}. 
We train them from scratch, and the test accuracy of each trained model is within 0.1\% of the best reported accuracy in their original papers. We generate the adversarial point clouds over each of them, and further evaluate the transferability of our proposed ITA attack among them.

\subsection{Experimental Settings}
\subsubsection{Evaluation Metrics}
To quantitatively evaluate the effectiveness of our proposed ITA attack and adversarial training defense, we measure by the {\it attack success rate}, which is the ratio of successfully fooling a 3D model. 
Besides, to measure the perturbation size of different attackers, we adopt four commonly used evaluation metrics: L2-norm distance $\mathcal{D}_{norm}$ \cite{cortes2012l2}, Chamfer distance $\mathcal{D}_{c}$ \cite{fan2017point}, Hausdorff distance $\mathcal{D}_{h}$ \cite{huttenlocher1993comparing}, as well as Geometric regularity $\mathcal{D}_{g}$ \cite{wen2020geometry}.
\red{Since the above metrics are all point-to-point distances, they fail to reveal the geometric fidelity of the point cloud.
In order to capture the imperceptibility well, we introduce the {\it geometric distortion metric} $\mathcal{D}_m$ in \cite{tian2017geometric}, which is a point-to-plane approach that integrates local plane properties and tracks visual qualities. 
Instead of measuring the original point-to-point error vectors directly as in previous distance metrics such as $\mathcal{D}_c$, $\mathcal{D}_m$ measures projected error vectors along normal directions, which would impose larger penalty on the errors that move further away from the local plane surface. 
For point clouds characterized by surfaces of structures, this metric is better aligned with the perceived quality than point-to-point metrics.
Specifically, given the benign point cloud $\bm{P}$ and its adversarial example $\bm{P}'$, $\mathcal{D}_m$ is defined as
\begin{equation}
    \mathcal{D}_m = \frac{1}{n} \sum_{\bm{p}_i' \in \bm{P}'} (\min_{\bm{p}_j \in \bm{P}} (\bm{p}_i' - \bm{p}_j) \cdot \bm{u}_{\bm{p}_j})^2,
\end{equation}
where $n$ is the number of points, $\bm{u}_{\bm{p}_j}$ is the unit normal vector of $\bm{p}_j$ to project the point-to-point distance along the normal direction.
}
We also report the attack success rates on several existing defenses (\eg, point removal \cite{zhou2019dup}, denoising \cite{wu2020if}) to further verify the superiority of our ITA attack compared with other attacks.

\subsubsection{Implementation Details}
To generate the adversarial examples, we adopt the Adam optimizer \cite{kingma2014adam} to optimize the objective of our proposed attack in Eq.~(\ref{eq:10}). We use a fixed learning schedule of 500 iterations, where the learning rate and momentum are set as 0.01 and 0.9, respectively. 
We set $k = 20$ to define local neighborhoods in Eq.~(\ref{eq:2}). 
We assign the weighting parameters $\lambda_1 = 0.1$ and $\lambda_2 = 1.0$ in Eq.~(\ref{eq:7}), (\ref{eq:8}) and (\ref{eq:10}).

To train the adversarial transformation model, we set the iteration numbers $L_1, L_2, L_3$ in Algorithm \ref{alg:adv} to 10, 500, 50, respectively. 
We determine the weighting parameters in Eq.~(\ref{eq:8}), (\ref{eq:9}) as $\alpha=1.0$, $\beta = 1.0$, and $\lambda_3=10$. 

During the adversarial training process of our defense method, we define the batch size as 64. 
We set $\omega_1=\omega_2=\omega_3=\omega_4=1.0$ in Eq.~(\ref{eq:11})(\ref{eq:12}). The weighting parameters $\omega_5,\omega_6$ in Eq.~(\ref{eq:13}) are set as 0.1.
All experiments are implemented on a single NVIDIA RTX 2080Ti GPU.

\begin{table}[t!]
    \centering
    \caption{Comparative results on the perturbation sizes of different methods required to achieve $100\%$ of attack success rate for adversarial point clouds.}
    \renewcommand{\arraystretch}{1.2}{
    \setlength{\tabcolsep}{1.4mm}{
    \begin{tabular}{|c|c|c|ccccc|}
    \hline
    Attack & \multirow{2}*{Methods} & Success & \multicolumn{5}{|c|}{Perturbation Size} \\ \cline{4-8}
    Model & ~ & Rate & $\mathcal{D}_{norm}$ & $\mathcal{D}_c$ & $\mathcal{D}_h$ & $\mathcal{D}_g$ & $\mathcal{D}_m$  \\ \hline
    \multirow{4}*{\rotatebox{90}{PointNet}} & FGSM & 100\% & 0.7936 & 0.1326 & 0.1853 & 0.3901 & \red{0.4667} \\
    ~ & 3D-ADV & 100\% & 0.3032 & \textbf{0.0003} & 0.0105 & 0.1772 & \red{0.1931}  \\
    ~ & GeoA & 100\% & 0.4385 & 0.0064 & 0.0175 & 0.0968 & \red{0.1243} \\
    ~ & Ours & 100\% & \textbf{0.1438} & 0.0037 & \textbf{0.0052} & \textbf{0.0671} & \red{\textbf{0.0927}} \\ \hline
    \multirow{4}*{\rotatebox{90}{PointNet++}} & FGSM & 100\% & 0.8357 & 0.1682 & 0.2275 & 0.4143 & \red{0.4512} \\
    ~ & 3D-ADV & 100\% & 0.3248 & \textbf{0.0005} & 0.0381 & 0.2034 & \red{0.2376} \\
    ~ & GeoA & 100\% & 0.4772 & 0.0198 & 0.0357 & 0.1141 & \red{0.1688} \\
    ~ & Ours & 100\% & \textbf{0.1465} & 0.0061 & \textbf{0.0076} & \textbf{0.0936} & \red{\textbf{0.1304}} \\ \hline
    \multirow{4}*{\rotatebox{90}{DGCNN}} & FGSM & 100\% & 0.8549 & 0.1890 & 0.2506 & 0.4217 & \red{0.4578}\\
    ~ & 3D-ADV & 100\% & 0.3326 & \textbf{0.0005} & 0.0475 & 0.2019 & \red{0.2486} \\
    ~ & GeoA & 100\% & 0.4933 & 0.0176 & 0.0402 & 0.1174 & \red{0.1735}\\
    ~ & Ours & 100\% & \textbf{0.1507} & 0.0058 & \textbf{0.0066} & \textbf{0.0913} & \red{\textbf{0.1489}} \\ \hline
    \red{\multirow{4}*{\rotatebox{90}{PointTrans.}}} & \red{FGSM} & \red{100\%} & \red{0.8332} & \red{0.1544} & \red{0.2379} & \red{0.4026} & \red{0.4457} \\
    ~ & \red{3D-ADV} & \red{100\%} & \red{0.3218} & \red{\textbf{0.0006}} & \red{0.0405} & \red{0.2012} & \red{0.2295} \\
    ~ & \red{GeoA} & \red{100\%} & \red{0.4837} & \red{0.0185} & \red{0.0383} & \red{0.1164} & \red{0.1736}\\
    ~ & \red{Ours} & \red{100\%} & \red{\textbf{0.1451}} & \red{0.0059} & \red{\textbf{0.0064}} & \red{\textbf{0.0887}} & \red{\textbf{0.1158}} \\ \hline
    \red{\multirow{4}*{\rotatebox{90}{PointMLP}}} & \red{FGSM} & \red{100\%} & \red{0.8029} & \red{0.1374} & \red{0.1948} & \red{0.3853} & \red{0.4384} \\
    ~ & \red{3D-ADV} & \red{100\%} & \red{0.3162} & \red{\textbf{0.0004}} & \red{0.0279} & \red{0.1895} & \red{0.1831} \\
    ~ & \red{GeoA} & \red{100\%} & \red{0.4578} & \red{0.0082} & \red{0.0235} & \red{0.0993} & \red{0.1265}\\
    ~ & \red{Ours} & \red{100\%} & \red{\textbf{0.1414}} & \red{0.0043} & \red{\textbf{0.0049}} & \red{\textbf{0.0706}} & \red{\textbf{0.0872}} \\\hline
    \end{tabular}}}
    \label{tab:imperceptibility}
\end{table}

\subsection{Evaluation on the Imperceptibility of Our ITA Attack}

\begin{figure*}[t!]
    \centering
    \includegraphics[width=1.0\textwidth]{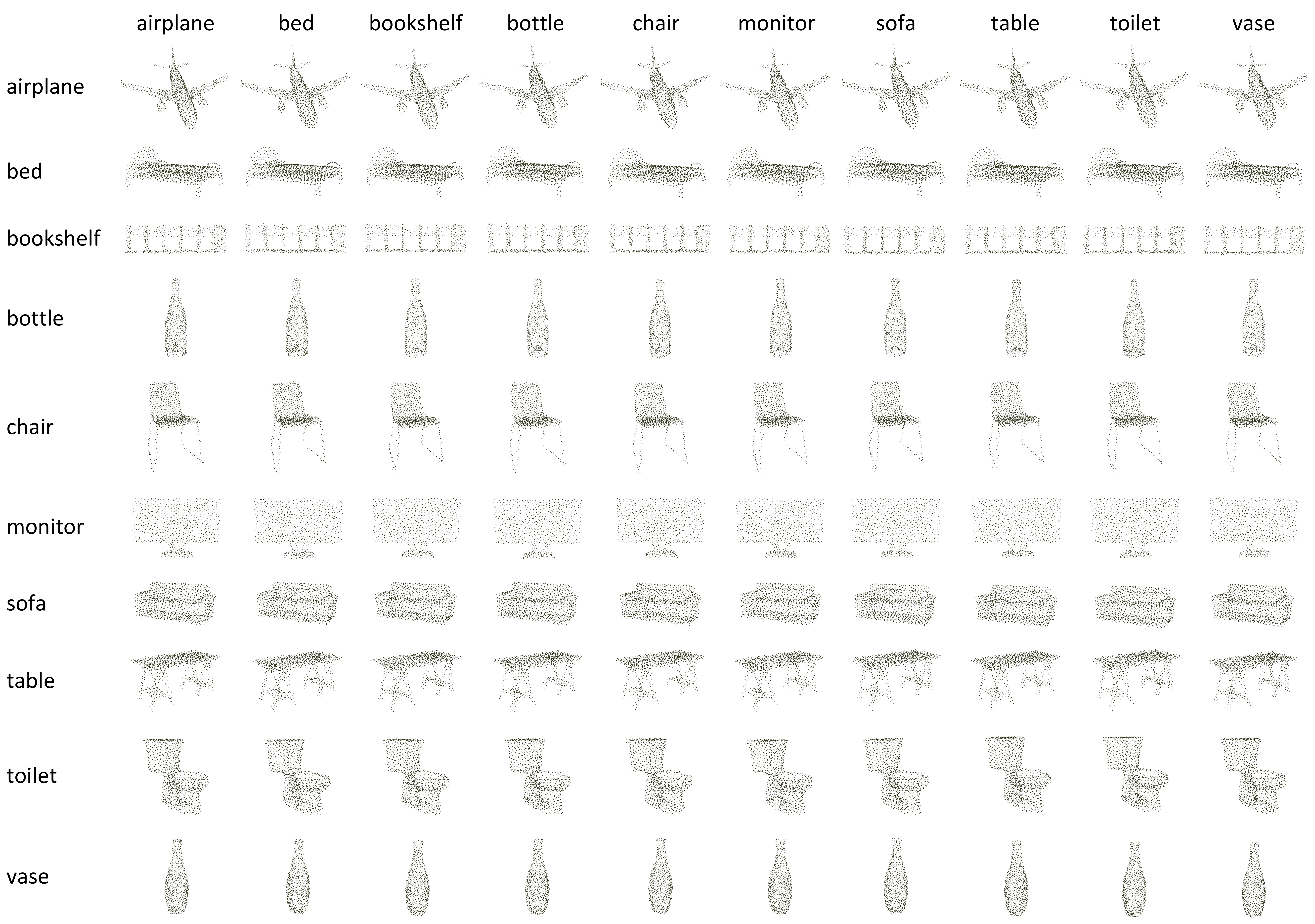}
    \caption{The generated adversarial point clouds from our proposed ITA attack. All the results are organized in a matrix form where the diagonal-entry instances present clean data belonging to a certain object class. The other off-diagonal-entry instances are the adversarial examples against the PointNet model, where the clean instance in each row is the input and the object class in each column is its attack target.}
    \label{fig:all_sample}
\end{figure*}

\begin{figure*}[t!]
    \centering
    \includegraphics[width=1.0\textwidth]{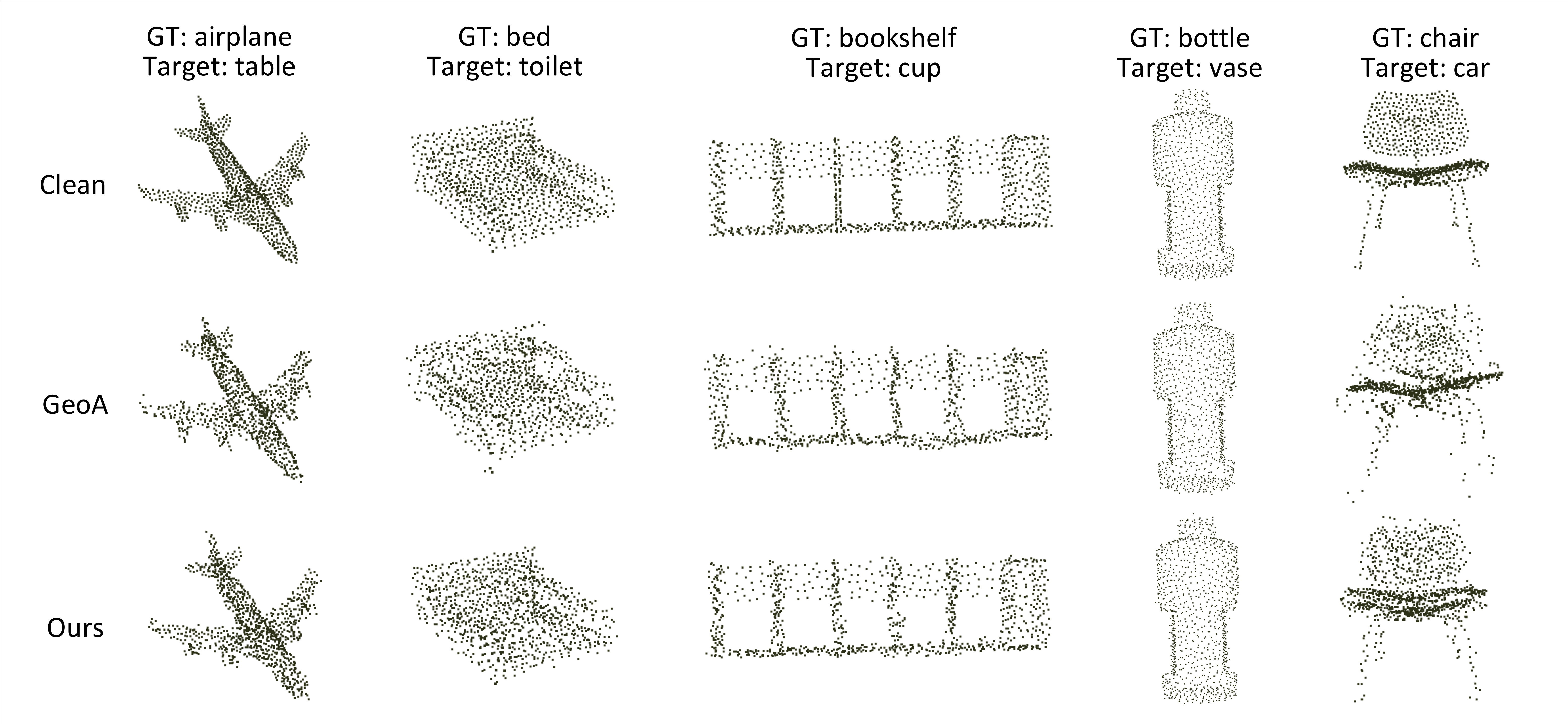}
    \caption{Adversarial point clouds of challenging cases from both GeoA and our proposed ITA attacks. Given a clean instance of each category, we generate its adversarial examples under two attacks for comparison.}
    \label{fig:challengs}
\end{figure*}

\begin{figure*}[t!]
    \centering
    \includegraphics[width=1.0\textwidth]{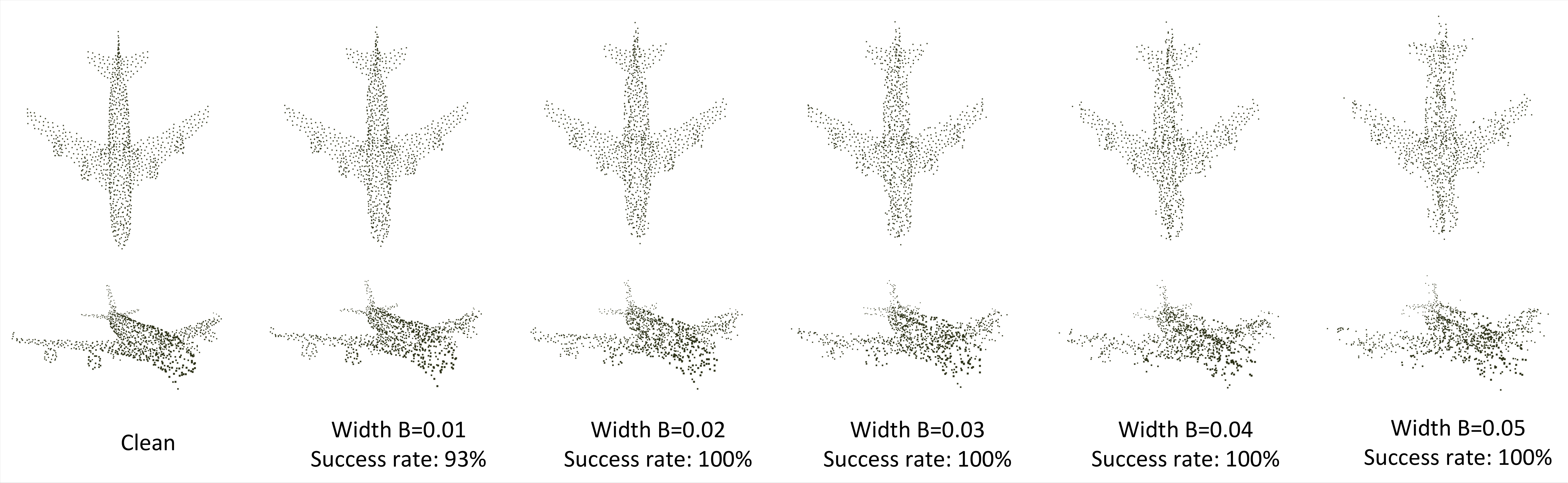}
    \caption{Sensitivity analysis on the bound width $B$. Experiments are conducted on the point clouds of 1024 points using PointNet.}
    \label{fig:width}
\end{figure*}

\subsubsection{Quantitative results}
To investigate the imperceptibility of our ITA attack, we measure the perturbation sizes of different adversarial point clouds required to achieve $100\%$ of attack success rate with \red{five} evaluation metrics. As shown in Table \ref{tab:imperceptibility}, our ITA attack generates adversarial point clouds with almost the lowest perturbation sizes in all evaluation metrics on \red{five} attack models. 
\red{Note that, 3D-ADV has the best Chamfer distance as the version of 3D-ADV that we compare to generates and places a set of independent points close to the original object, without perturbing existing points in the point cloud. 
However, it induces much larger distortions than ours on the other four metrics.}
In particular, we require the lowest perturbation size in terms of the geometric regularity $\mathcal{D}_g$ and the geometric distortion metric $\mathcal{D}_m$, which indicates that our generated adversarial point clouds reveal the geometric fidelity best. 
\red{Compared to the GeoA which optimizes a specific curvature loss with xyz-directional perturbations to keep the original geometry information, our attack constrains the perturbation direction along the normal vector to directly keep the position dependency among neighboring points, thus promoting the consistency of local curvatures and achieving smaller $\mathcal{D}_g$.}
Besides, for each attack method, it takes larger perturbation sizes to successfully attack \red{PointNet++, DGCNN and PointTrans.} than to attack PointNet and \red{PointMLP}, which indicates that PointNet++, DGCNN and PointTrans. are harder to attack.  

\subsubsection{Visualization results}
We also show the visualization results of the generated adversarial examples in Figure \ref{fig:all_sample}. Here, an example instance of each category is attacked against the PointNet model targeting at all the other 9 categories. 
We observe that, all adversarial point clouds exhibit similar geometric structures with their corresponding benign point clouds, \ie, the attacks are very imperceptible to humans. Besides, our adversarial examples have no outliers or uneven point distributions in local area.

Furthermore, to provide detailed comparisons over the quality of the generated adversarial examples with the state-of-the-art attack method GeoA \cite{wen2020geometry}, we demonstrate some visualization results of most challenging attack cases in Figure \ref{fig:challengs}, where five example instances of different categories are attacked against PointNet with their challenging attack targets. 
We observe that the adversarial examples generated by GeoA destruct the local point distribution in almost all cases (\ie, airplane, bed, bookshelf, and chair), which are easily noticeable by humans. 
Also, GeoA exhibits a few outliers (\ie, airplane, bed, chair), which can be detected by outlier-removal-based defense methods. 
In contrast, our adversarial point clouds have no outliers, and even have no uneven local distribution, leading to significantly higher imperceptibility. 
This is because our ITA not only perturbs each point along the normal direction, but also constrains them with a strict width, which will be analyzed below.


\subsubsection{Sensitivity analysis on bound width $B$}
\label{exp:bound}
We further investigate whether adversarial effects vary with respect to different settings of the bound width $B$, as shown in Figure~\ref{fig:width}. 
In this figure, when perturbing the clean point cloud with bound width $B=0.01$, the generated adversarial example has almost the same geometric structure as the clean one and is perfectly imperceptible to humans. 
However, a tight bound limits the perturbation degree of points, which results in only $93\%$ of success rate of all generated adversarial examples. With the gradual increase of the bound width $B$, the success rate of our ITA attack also increases to $100\%$, but the adversarial point cloud becomes more messy and perceptible. 
To strike a good trade-off between the imperceptibility of the generated point cloud and the success rate of our ITA attack, we empirically set $B=0.02$ in all experiments.

\begin{table}[t!]
    \centering
    \caption{\red{Comparative results on the attack success rate of different attack methods. Experiments are implemented on PointNet model.}}
    \renewcommand{\arraystretch}{1.2}{
    \setlength{\tabcolsep}{2.0mm}{
    \begin{tabular}{|c|ccccc|}
    \hline
    \multirow{2}*{\red{Methods}} & \multicolumn{5}{c|}{\red{Budget $B$}} \\ \cline{2-6}
    ~ & \red{0.01} & \red{0.02} & \red{0.03} & \red{0.04} & \red{0.05}  \\ \hline
    \red{FGSM} & \red{42\%} & \red{67\%} & \red{82\%} & \red{91\%} & \red{98\%} \\
    \red{3D-ADV} & \red{13\%} & \red{24\%} & \red{45\%} & \red{61\%} & \red{73\%} \\
    \red{GeoA} & \red{79\%} & \red{88\%} & \red{99\%} & \red{\textbf{100}}\% & \red{\textbf{100}\% } \\
    \red{Ours} & \red{\textbf{93}\%} & \red{\textbf{100}\%} & \red{\textbf{100}\%} & \red{\textbf{100}\%} & \red{\textbf{100}\%}  \\ \hline
    \end{tabular}}}
    \label{tab:bandd1}
\end{table}

\begin{table}[t!]
    \centering
    \caption{\red{Comparative results on the perturbation sizes of different methods required to achieve $100\%$ of attack success rate for adversarial point clouds without budget $B$. Experiments are implemented on PointNet.}}
    \renewcommand{\arraystretch}{1.2}{
    \setlength{\tabcolsep}{1.4mm}{
    \begin{tabular}{|c|c|ccccc|}
    \hline
    \multirow{2}*{\red{Methods}} & \red{Success} & \multicolumn{5}{c|}{\red{Perturbation Size}} \\ \cline{3-7}
    ~ & \red{Rate} & \red{$\mathcal{D}_{norm}$} & \red{$\mathcal{D}_c$} & \red{$\mathcal{D}_h$} & \red{$\mathcal{D}_g$} & \red{$\mathcal{D}_m$}  \\ \hline
    \red{FGSM} & \red{100\%} & \red{0.7936} & \red{0.1326} & \red{0.1853} & \red{0.3901} & \red{0.4667} \\
    \red{3D-ADV} & \red{100\%} & \red{0.3032} & \red{\textbf{0.0003}} & \red{0.0105} & \red{0.1772} & \red{0.1931} \\
    \red{GeoA} & \red{100\%} & \red{0.4385} & \red{0.0064} & \red{0.0175} & \red{0.0968} & \red{0.1243} \\
    \red{Ours} & \red{100\%} & \red{\textbf{0.1716}} & \red{0.0053} & \red{\textbf{0.0087}} & \red{\textbf{0.0784}} & \red{\textbf{0.1018}} \\ \hline
    \end{tabular}}}
    \label{tab:bandd2}
\end{table}

\begin{table*}[t!]
    \centering
    \caption{The transfer-based attack success rates (\%) on \red{five} models by various attacks {\it without} our proposed adversarial transformation model. The white-box target models are adopted to generate
    adversarial examples, while the black-box victim models are the attack targets.}
    \renewcommand{\arraystretch}{1.2}{
    \setlength{\tabcolsep}{2.0mm}{
    \begin{tabular}{|c|c|ccccc|}
    \hline
    White-box & Attack & \multicolumn{5}{c|}{Black-box Victim Model} \\ \cline{3-7}
    Target Model & Methods & PointNet & PointNet++ & DGCNN & PointTrans. & PointMLP \\ \hline
    \multirow{4}*{PointNet} & FGSM & \textbf{100.0} & 3.99 & 0.63 & \red{0.57} & \red{1.68} \\
    ~ & 3D-ADV & \textbf{100.0} & 8.45 & 1.28 & \red{1.13} & \red{3.96} \\
    ~ & GeoA & \textbf{100.0} & 11.59 & 2.59 & \red{1.97} & \red{5.74} \\
    ~ & Ours & \textbf{100.0} & \textbf{14.60} & \textbf{4.43} & \red{\textbf{4.15}} & \red{\textbf{9.31}} \\ \hline
    \multirow{4}*{PointNet++} & FGSM & 3.16 & \textbf{100.0} & 5.57 & \red{4.63} & \red{2.89} \\
    ~ & 3D-ADV & 6.63 & \textbf{100.0} & 10.98 & \red{10.12} & \red{5.74} \\
    ~ & GeoA & 9.47 & \textbf{100.0} & 19.77 & \red{16.81} & \red{7.38} \\
    ~ & Ours & \textbf{11.68} & \textbf{100.0} & \textbf{23.21} & \red{\textbf{21.95}} & \red{\textbf{10.03}} \\ \hline
    \multirow{4}*{DGCNN} & FGSM & 3.59 & 7.21 & \textbf{100.0} & \red{5.79} & \red{3.17} \\
    ~ & 3D-ADV & 6.82 & 13.53 & \textbf{100.0} & \red{11.84} & \red{6.26} \\
    ~ & GeoA & 12.46 & 24.24 & \textbf{100.0} & \red{19.40} & \red{10.15} \\
    ~ & Ours & \textbf{18.71} & \textbf{29.89} & \textbf{100.0} & \red{\textbf{25.93}} & \red{\textbf{15.78}} \\ \hline
    \red{\multirow{4}*{PointTrans.}} & \red{FGSM} & \red{3.25} & \red{7.38} & \red{6.43} & \red{\textbf{100.0}} & \red{3.51} \\
    ~ & \red{3D-ADV} & \red{6.54} & \red{13.60} & \red{13.02} & \red{\textbf{100.0}} & \red{6.89} \\
    ~ & \red{GeoA} & \red{12.57} & \red{23.64} & \red{22.15} & \red{\textbf{100.0}} & \red{12.29} \\
    ~ & \red{Ours} & \red{\textbf{19.28}} & \red{\textbf{29.46}} & \red{\textbf{27.73}} & \red{\textbf{100.0}} & \red{\textbf{18.16}} \\ \hline
    \red{\multirow{4}*{PointMLP}} & \red{FGSM} & \red{4.35} & \red{3.81} & \red{1.74} & \red{1.09} & \red{\textbf{100.0}} \\
    ~ & \red{3D-ADV} & \red{9.27} & \red{8.60} & \red{4.52} & \red{3.54} & \red{\textbf{100.0}} \\
    ~ & \red{GeoA} & \red{12.38} & \red{11.63} & \red{6.75} & \red{5.72} & \red{\textbf{100.0}} \\
    ~ & \red{Ours} & \red{\textbf{16.94}} & \red{\textbf{16.31}} & \red{\textbf{10.46}} & \red{\textbf{9.83}} & \red{\textbf{100.0}} \\ \hline
    \end{tabular}}}
    \label{tab:transfer1}
\end{table*}

\red{Besides, we also investigate how other attacks perform during the adversarial example generation with the bound $B$. We show the experimental results in Table~\ref{tab:bandd1} where all attack models are constrained by the same budget $B$. 
Specifically, for attack models that perturb points along xyz directions, $B$ denotes the radius of the ball of adversarial space for each point.
It shows that our attack still achieves almost 100\% success rate and outperforms other methods significantly especially under a small budget.
This demonstrates that perturbing along the normal vector is a geometry-aware perturbation direction while it is hard for other attacks to search their desired perturbation directions. 
Moreover, we {\it remove} the bound constraint during the adversarial example generation and directly evaluate the perturbation sizes required to achieve 100\% success rate. As shown in Table~\ref{tab:bandd2}, even without the bound constraint, our attack still requires the lowest perturbation sizes. 
Other methods are more perceptible than ours when achieving the same attack success rate.}


\subsection{Evaluation on the Transferability of Our ITA Attack}
\label{exp:transfer}


\begin{table*}[t!]
    \centering
    \caption{The transfer-based attack success rates (\%) on \red{five} models by various attacks {\it with} our proposed \red{adversarial} adversarial transformation model. The white-box target models are adopted to generate adversarial examples, while the black-box victim models are the attack targets.}
    \renewcommand{\arraystretch}{1.2}{
    \setlength{\tabcolsep}{2.0mm}{
    \begin{tabular}{|c|c|ccccc|}
    \hline
    White-box & Attack & \multicolumn{5}{c|}{Black-box Victim Model} \\ \cline{3-7}
    Target Model & Methods & PointNet & PointNet++ & DGCNN & PointTrans. & PointMLP \\ \hline
    \multirow{4}*{PointNet} & FGSM & \textbf{100.0} & 38.27 & 34.52 & \red{32.79} & \red{36.40} \\
    ~ & 3D-ADV & \textbf{100.0} & 43.41 & 35.75 & \red{33.42} & \red{38.64} \\
    ~ & GeoA & \textbf{100.0} & 49.33 & 40.83 & \red{38.53} & \red{44.29} \\
    ~ & Ours & \textbf{100.0} & \textbf{54.62} & \textbf{42.19} & \red{\textbf{41.38}} & \red{\textbf{48.91}} \\ \hline
    \multirow{4}*{PointNet++} & FGSM & 28.71 & \textbf{100.0} & 47.84 & \red{39.04} & \red{27.37}\\
    ~ & 3D-ADV & 32.38 & \textbf{100.0} & 52.49 & \red{45.89} & \red{30.77} \\
    ~ & GeoA & 37.57 & \textbf{100.0} & 58.02 & \red{52.65} & \red{35.28}  \\
    ~ & Ours & \textbf{42.43} & \textbf{100.0} & \textbf{62.38} & \red{\textbf{61.12}} & \red{\textbf{40.41}} \\ \hline
    \multirow{4}*{DGCNN} & FGSM & 45.58 & 54.90 & \textbf{100.0} & \red{49.44} & \red{43.79} \\
    ~ & 3D-ADV & 54.77 & 62.51 & \textbf{100.0} & \red{59.93} & \red{53.18} \\
    ~ & GeoA & 62.33 & 69.98 & \textbf{100.0} & \red{66.35} & \red{58.27} \\
    ~ & Ours & \textbf{67.06} & \textbf{76.74} & \textbf{100.0} & \red{\textbf{73.50}} & \red{\textbf{62.09}} \\ \hline
    \red{\multirow{4}*{PointTrans.}} & \red{FGSM} & \red{44.73} & \red{55.38} & \red{51.06} & \red{\textbf{100.0}} & \red{45.11} \\
    ~ & \red{3D-ADV} & \red{54.58} & \red{63.07} & \red{61.24} & \red{\textbf{100.0}} & \red{53.90} \\
    ~ & \red{GeoA} & \red{63.46} & \red{70.31} & \red{68.65} & \red{\textbf{100.0}} & \red{59.92} \\
    ~ & \red{Ours} & \red{\textbf{68.69}} & \red{\textbf{76.53}} & \red{\textbf{75.37}} & \red{\textbf{100.0}} & \red{\textbf{62.14}} \\ \hline
    \red{\multirow{4}*{PointMLP}} & \red{FGSM} & \red{40.47} & \red{38.65} & \red{37.94} & \red{37.38} & \red{\textbf{100.0}} \\
    ~ & \red{3D-ADV} & \red{46.29} & \red{44.73} & \red{41.20} & \red{40.61} & \red{\textbf{100.0}} \\
    ~ & \red{GeoA} & \red{51.52} & \red{50.14} & \red{45.35} & \red{44.47} & \red{\textbf{100.0}} \\
    ~ & \red{Ours} & \red{\textbf{58.79}} & \red{\textbf{58.08}} & \red{\textbf{51.23}} & \red{\textbf{49.86}} & \red{\textbf{100.0}} \\ \hline
    \end{tabular}}}
    \label{tab:transfer3}
\end{table*}

\subsubsection{Without the \red{adversarial} transformation model}
We first perform four adversarial attacks, namely FGSM \cite{zhang2019adversarial}, 3D-ADV \cite{xiang2019generating}, GeoA \cite{wen2020geometry}, and our proposed ITA, on a single white-box 3D model without our proposed adversarial transformation model. We craft adversarial examples on normally trained models and test them on all the \red{five} models we consider. 
The corresponding transfer-based attack success rates are shown in Table \ref{tab:transfer1}. 
The rows and columns of the table present the models we attack and the \red{five} models we test. 
We observe that although all the attacks achieve very low success rates for transfer-based attacks, our ITA still has relatively higher success rates than others. This is because our directional perturbation strategy retains the geometric properties of the clean point clouds well, without outliers or uneven local distribution, which is thus more robust.

\red{The above results also indicate the connection between the imperceptibility and transferability:
To achieve the imperceptibility to humans, an adversarial point cloud should keep the geometric characteristics, \ie, the surface structure of the underlying manifold. 
That is, an imperceptible attack keeps the dependency among neighboring points so as to preserve the geometric contexts of benign point clouds, such as slowly-varying smooth surfaces and contours with large variations. 
Such crafted adversarial examples tend to be indistinguishable to different network models that generally learn the geometry of 3D shapes for classification, thus enhancing the transferability.}



\subsubsection{With the \red{adversarial} transformation model}

The success rates of transfer-based attacks with the linear-based adversarial transformation model are listed in Table \ref{tab:transfer3}. 
Compared to the transfer-based attack results in Table \ref{tab:transfer1}, all the attacks outperform the previous two variants by a large margin on all the black-box models. 
Specifically, the ITA attack with \red{adversarial} transformation improves the transfer success rate by
\red{40.02\%, 37.76\%, 37.23\% and 39.60\% on PointNet++, DGCNN, PointTrans. and PointMLP models,} respectively. 
The main reason is that, \red{the proposed adversarial transformation model learns a combination of multiple simple transformations that infers the most harmful deformations to the adversarial examples. During the adversarial learning, the generated adversarial examples are able to resist such distortions, thus they are transferable to attack black-box models with higher success rates.}

\subsubsection{\red{Comparison with Learning-free Transformations}}
\red{Since our proposed adversarial transformation module is learning-based and requires the adversarial learning process, we also conduct experiments to replace this module with various learning-free analytic transformations for validating the benefits of learning-based transformation modules compared to the learning-free ones. 
In particular, we add experiments on directly improving the transferability of adversarial samples by training them to be robust against common learning-free transformations, including cloud-wise and point-wise learning-free transformations.}

\begin{table}[t!]
    \centering
    \caption{\red{Comparison on our ITA attack {\it with} learning-based (\ie, Linear-based) and learning-free transformation. ``Learning-free$^c$" denotes the cloud-wise learning-free transformation, and ``Learning-free$^p$" denotes the point-wise one.}}
    \renewcommand{\arraystretch}{1.2}{
    \setlength{\tabcolsep}{1.4mm}{
    \begin{tabular}{|c|c|ccc|}
    \hline
    \red{White-box} & \red{Transformation} & \multicolumn{3}{c|}{\red{Black-box Victim Model}} \\ \cline{3-5}
    \red{Target Model} & \red{Methods} & \red{PointNet} & \red{PointNet++} & \red{DGCNN} \\ \hline
    \red{\multirow{4}*{PointNet}} & \red{None} & \red{\textbf{100.0}} & \red{14.60} & \red{4.43} \\
    ~ & \red{Learning-free$^c$} & \red{\textbf{100.0}} & \red{31.28} & \red{20.04} \\
    ~ & \red{Learning-free$^p$} & \red{\textbf{100.0}} & \red{38.41} & \red{26.76} \\
    ~ & \red{Learning-based} & \red{\textbf{100.0}} & \red{\textbf{54.62}} & \red{\textbf{42.19}} \\ \hline
    \red{\multirow{4}*{PointNet++}} & \red{None} & \red{11.68} & \red{\textbf{100.0}} & \red{20.21} \\
    ~ & \red{Learning-free$^c$} & \red{22.37} & \red{\textbf{100.0}} & \red{37.98} \\
    ~ & \red{Learning-free$^p$} & \red{28.75} & \red{\textbf{100.0}} & \red{44.12}\\
    ~ & \red{Learning-based} & \red{\textbf{42.43}} & \red{\textbf{100.0}} & \red{\textbf{62.38}} \\ \hline
    \red{\multirow{4}*{DGCNN}} & \red{None} & \red{18.71} & \red{29.89} & \red{\textbf{100.0}} \\
    ~ & \red{Learning-free$^c$} & \red{36.85} & \red{46.20} & \red{\textbf{100.0}} \\
    ~ & \red{Learning-free$^p$} & \red{42.57} & \red{51.92} & \red{\textbf{100.0}}\\
    ~ & \red{Learning-based} & \red{\textbf{67.06}} & \red{\textbf{76.74}} & \red{\textbf{100.0}} \\ \hline
    \end{tabular}}}
    \label{tab:transfer}
\end{table}

\red{We
adopt four types of cloud-wise/point-wise learning-free transformations:
1) Translation: randomly translate the x-, y-, z-coordinates of the whole point cloud by three parameters in the range $[-0.2, 0.2]$ or of each point cloud by three parameters in the range $[-0.02, 0.02]$. 
2) Rotation: randomly rotate the point cloud by an angle in the range $[-180^{\circ}, 180^{\circ}]$ or rotate each point by an angle in the range $[-5^{\circ}, 5^{\circ}]$ along the gravity axis. 
3) Shearing: randomly shear the x-, y-, z-coordinates of the whole point cloud with the six parameters of a shearing matrix in the range $[-0.2, 0.2]$ or shear those of each point in $[-0.1, 0.1]$. 
4) Jittering: randomly generate jittering on the whole point cloud from a Gaussian distribution $\mathcal{N}(0, 0.01^2)$ with the output clipped at 0.05 or on each point clipped at 0.02.}


\red{As shown in Table~\ref{tab:transfer}, the point-wise learning-free transformation model performs better than the cloud-wise one in terms of the transferability, since the former tends to generate more harmful distortions.
Moreover, our learning-based transformation model performs much better than the learning-free ones. 
This is because the learning-free transformation models are limited by the random settings of different specific transformations, while the learning-based one is able to adaptively learn to construct the most harmful distortion on the point clouds.}

\begin{figure}[t!]
    \centering
    \includegraphics[width=0.48\textwidth]{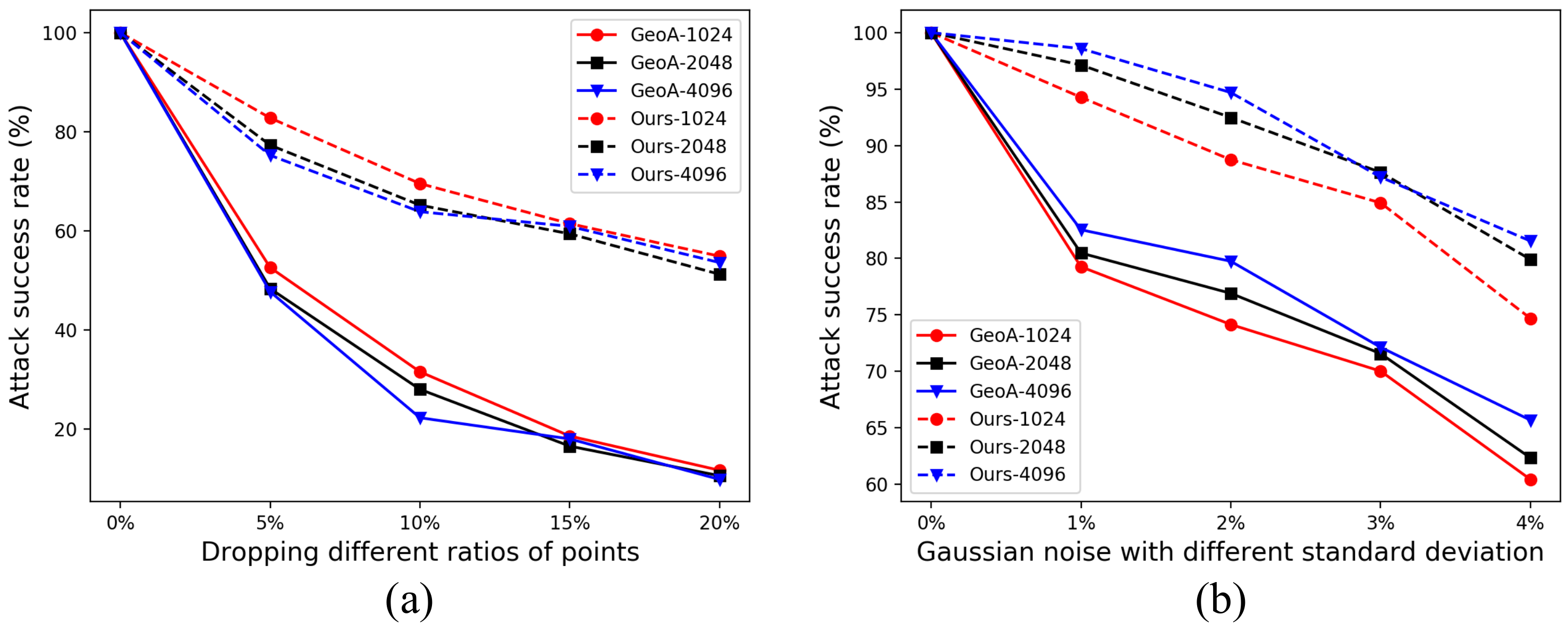}
    \caption{Robustness of our proposed ITA under (a) missing data by dropping points and (b) Gaussian noise. The success rates (\%) are reported over attacking PointNet model. We also sample different numbers of points (1024, 2048, or 4096) from the original objects for detailed comparisons.}
    \label{fig:noise}
\end{figure}

\subsection{Evaluation on Existing Point Cloud Defense}
\subsubsection{Robustness to Noise or Incompleteness}
To further examine the robustness of our proposed ITA attack, we imitate the real-world scenarios that point clouds may suffer from noise or missing data. Therefore, 
we conduct such experiments by dropping 0\%, 5\%, 10\%, 15\%, 20\% of points respectively from the adversarial point cloud or perturbing it by Gaussian noise with standard deviation from 0\% to 4\% of the bounding sphere’s radius. 
Then, we measure their attack success rates on the same 3D model.
Meanwhile, we investigate the effect of the number of points in the point cloud. As shown in Figure \ref{fig:noise}(a), we see that while dropping more points will generally result in lower attack success rates, the performance of our ITA attack decays much slower than that of GeoA, which validates the robustness of our model. We also observe that sampled points with a smaller number are more robust to point dropping. 
In addition, as shown in Figure \ref{fig:noise}(b), the ratio of the standard deviation of the added noise also has impact on the attack success rates of adversarial point clouds, where our method achieves much higher success rates than GeoA over all noise levels. We also observe that sampled points of a larger number are more robust to the Gaussian noise, because more points ensure a higher tolerance for random noise.
In summary, our ITA attack is significantly more robust than the GeoA attack in the presence of noisy or incomplete point cloud data.

\red{Besides, we investigate whether randomly perturbing each point along its normal direction would break our attack. Specifically, we perturb our adversarial examples by adding Gaussian noise with standard deviation from 0\% to 4\% of the bounding sphere’s radius along the normal vector of each point. As presented in Table~\ref{tab:defense_new}, our adversarial examples are quite robust to such simple defense strategy, demonstrating the strength of the proposed ITA attack.}

\begin{table}[t!]
    \centering
    \caption{\red{Defense by adding different ratios of Gaussian noise along the normal direction.}}
    \setlength{\tabcolsep}{3.5mm}{
    \begin{tabular}{c|ccccc}
    \hline
    \red{\multirow{3}*{Methods}} & \multicolumn{5}{c}{\red{Attack success rate (\%)}} \\
    ~ & \multicolumn{5}{c}{\red{adding different ratios of Gaussian noise}} \\ \cline{2-6}
    ~ & \red{0\%} & \red{0.5\%} & \red{1\%} & \red{1.5\%} & \red{2\%} \\ \hline
    \red{Ours} & \red{100} & \red{100} & \red{99.24} & \red{97.88} & \red{95.72} \\ \hline
    \end{tabular}}
    \label{tab:defense_new}
\end{table}

\begin{table}[t!]
    \centering
    \caption{The error rates (\%) on PointNet by various attacks under various \red{existing} defense methods.}
    \renewcommand{\arraystretch}{1.4}{
    \begin{tabular}{|c|c|c|c|c|c|}
    \hline
    Attack & \red{None} & FGSM & 3D-ADV & GeoA & Ours \\ \hline
    No Defense & \red{0.00} & \textbf{100.0} & \textbf{100.0} & \textbf{100.0} & \textbf{100.0} \\ \hline
    SRS & \red{0.93} & 9.68 & 22.53 & 67.61 & \textbf{91.85} \\ \hline
    SOR & \red{0.58} & 6.26 & 17.19 & 62.47 & \textbf{90.37} \\ \hline
    DUP-Net & \red{0.64} & 4.38 & 15.44 & 59.15 & \textbf{85.41}\\ \hline
    IF-Defense & \red{0.67} & 4.80 & 13.70 & 38.72 & \textbf{69.32} \\ \hline
    \end{tabular}}
    \label{tab:defense}
\end{table}

\subsubsection{Attacks under existing defenses}
\label{exp:defense}
Firstly, we investigate whether our ITA attack is still effective when attacking the white-box 3D model with existing defense methods.
That is, we aim to verify our ITA attack under various defenses.
To evaluate the adversarial robustness of various attacks, we employ the following defense methods: Simple Random Sampling (SRS) \cite{zhang2019adversarial}, Statistical Outlier Removal (SOR) \cite{zhou2019dup}, DUP-Net defense \cite{zhou2019dup} and IF-Defense \cite{wu2020if}.
We present the success rates by various attacks under each defense method in Table \ref{tab:defense}, and have the following observations:
\red{(1) the error rate on clean data is equal or close to 0 perhaps due to the modification of point clouds by the defense methods (such as sampling, point removal, \etc).}
(2) FGSM and 3D-ADV attacks have low success rates under all the defenses. 
This is because that FGSM shifts points along xyz directions without any constraint of the length, which easily results in uneven local distribution and outliers. 
Meanwhile, 3D-ADV adds perturbed points/clusters/objects, which is also easily detectable by the defenses.
(3) GeoA achieves higher attack success rates than FGSM and 3D-ADV under various defenses, since it utilizes a geometry-aware loss function to constrain the similarity of curvatures in closest point pairs and thus leads to fewer outliers. 
However, GeoA still cannot effectively attack against the IF-Defense, because GeoA also destructs the local distribution by perturbing points in xyz directions.
(4) Our ITA attack achieves the highest success rates than all other attacks under all defenses. 
Since we choose the point-wise normal vector as the perturbation direction and constrain the shifting length within a strictly bounded width, our adversarial examples have no outliers and alleviate the change in the local distribution. Hence, we achieve much better attack performance especially under the IF-Defense.

\begin{table}[t!]
    \centering
    \caption{The transfer-based attack success rates (\%) on PointNet by the proposed ITA attack with our \red{adversarial} transformation model. * indicates the black-box victim model adopts the defense method.}
    \renewcommand{\arraystretch}{1.2}{
    \setlength{\tabcolsep}{1.4mm}{
    \begin{tabular}{|c|c|ccc|}
    \hline
    White-box & Attack & \multicolumn{3}{c|}{Defense: Ours} \\ \cline{3-5}
    Target Model & Methods & PointNet & PointNet++* & DGCNN* \\ \hline
    \multirow{4}*{PointNet} & FGSM & \red{8.65} & \red{3.54} & \red{0.58} \\
    ~ & 3D-ADV & \red{12.84} & \red{6.02} & \red{1.17}\\
    ~ & GeoA & \red{15.72} & \red{8.36} & \red{2.41}\\
    ~ & Ours & \red{18.13} & \red{9.89} & \red{4.27} \\ \hline \hline
     White-box & Attack & \multicolumn{3}{c|}{\red{Defense: Vanilla adversarial training \cite{liu2019extending}}} \\ \cline{3-5}
    Target Model & Methods & PointNet & PointNet++* & DGCNN* \\ \hline
    \multirow{4}*{PointNet} & FGSM & \red{31.72} & \red{20.88} & \red{17.39} \\
    ~ & 3D-ADV & \red{39.14} & \red{23.85} & \red{18.47}\\
    ~ & GeoA & \red{44.26} & \red{27.50} & \red{22.11}\\
    ~ & Ours & \red{49.98} & \red{32.78} & \red{24.93} \\ \hline \hline
     \red{White-box} & \red{Attack} & \multicolumn{3}{c|}{\red{Defense: Adversarial training with \cite{sun2020adversarial}}} \\ \cline{3-5}
    \red{Target Model} & \red{Methods} & \red{PointNet} & \red{PointNet++*} & \red{DGCNN*} \\ \hline
    \red{\multirow{4}*{PointNet}} & \red{FGSM} & \red{21.92} & \red{14.97} & \red{9.22} \\
    ~ & \red{3D-ADV} & \red{29.05} & \red{17.21} & \red{13.02}\\
    ~ & \red{GeoA} & \red{32.86} & \red{20.49} & \red{15.37}\\
    ~ & \red{Ours} & \red{37.38} & \red{24.79} & \red{17.88} \\ \hline
    \end{tabular}}}
    \label{tab:defense1}
\end{table}

\begin{figure*}[t!]
    \centering
    \includegraphics[width=1.0\textwidth]{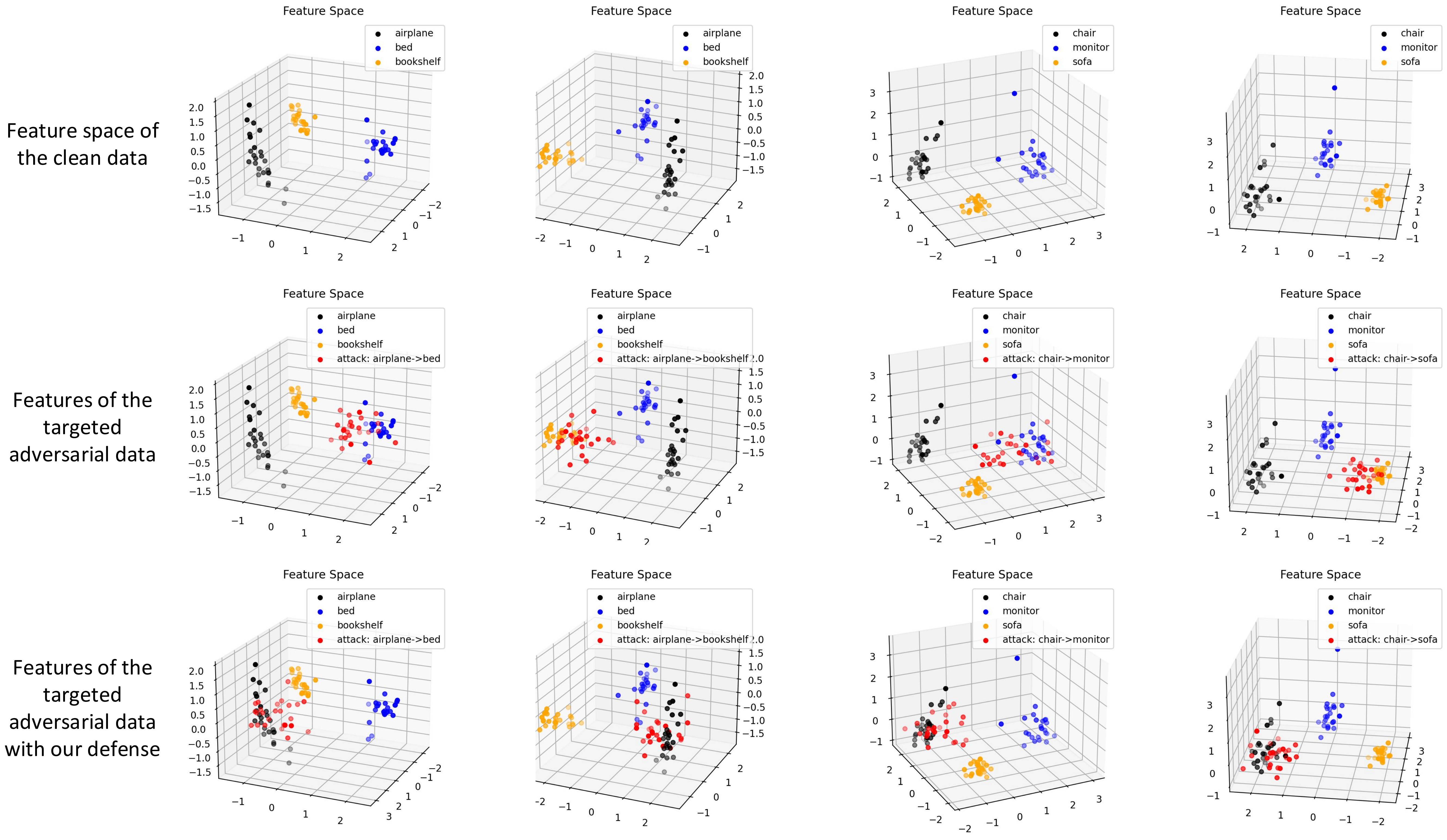}
    \caption{Visualization examples on feature distribution of clean and adversarial point clouds in the latent space. The second and third rows represent the changes of feature distribution of the adversarial examples (red nodes) without or with our defense, respectively. 
    In each sub-figure, we sample 25 benign point clouds for each object class and utilize Principal Component Analysis (PCA) to reduce their global features into 3 dimensions. We plot each point cloud with its 3-dimensional feature as a node, and distinguish them by different colors (black, yellow, blue or red). The red nodes represent adversarial point clouds generated from the black nodes which target at the blue or orange nodes.}
    \label{fig:feature}
\end{figure*}

\subsection{Evaluation of Our Proposed Defense}
To verify the effectiveness of our proposed adversarial training based defense method, we first train the victim models from scratch with our defense method, and then measure \red{both the white-box and} the transfer-based attack success rates on them.
As shown in Table \ref{tab:defense1}, \ref{tab:defense2} and \ref{tab:defense3}, we evaluate the effectiveness of our defense method on three commonly adopted 3D models, respectively. 
Specifically, \red{we compare our defense method (adversarial training + constraint loss in the latent space) with two adversarial training based methods: 
1) Vanilla adversarial training \cite{liu2019extending}, which simply adopts the adversarial training strategy without adopting our proposed inter-class and intra-class losses in the latent space. 2) Adversarial training with \cite{sun2020adversarial}, which replaces the pooling layer of the victim model with its proposed "DeepSym" and then utilizes the adversarial training. 
Note that, since \cite{sun2020adversarial} claims that their approach is applicable on the PointNet model but not on the PointNet++ or DGCNN, we only implement this defense on PointNet.}

\begin{table}[t!]
    \centering
    \caption{The transfer-based attack success rates (\%) on PointNet++ by the proposed ITA attack with our \red{adversarial} transformation model. * indicates the black-box victim model adopts the defense method.}
    \renewcommand{\arraystretch}{1.2}{
    \setlength{\tabcolsep}{1.4mm}{
    \begin{tabular}{|c|c|ccc|}
    \hline
    White-box & Attack & \multicolumn{3}{c|}{Defense: Ours} \\ \cline{3-5}
    Target Model & Methods & PointNet* & PointNet++ & DGCNN* \\ \hline
    \multirow{4}*{PointNet++} & FGSM & \red{2.44} & \red{10.74} & \red{4.37} \\
    ~ & 3D-ADV & \red{4.92} & \red{13.88} & \red{7.15}\\
    ~ & GeoA & \red{7.01} & \red{18.63} & \red{11.02}\\
    ~ & Ours & \red{9.23} & \red{22.28} & \red{12.85} \\ \hline \hline
     White-box & Attack & \multicolumn{3}{c|}{\red{Defense: Vanilla adversarial training \cite{liu2019extending}}} \\ \cline{3-5}
    Target Model & Methods & PointNet* & PointNet++ & DGCNN* \\ \hline
    \multirow{4}*{PointNet++} & FGSM & \red{17.85} & \red{37.64} & \red{28.91}\\
    ~ & 3D-ADV & \red{20.46} & \red{42.75} & \red{31.37}\\
    ~ & GeoA & \red{23.21} & \red{49.93} & \red{37.59}\\
    ~ & Ours & \red{27.38} & \red{55.19} & \red{41.88} \\ \hline
    \end{tabular}}}
    \label{tab:defense2}
\end{table}

\begin{table}[t!]
    \centering
    \caption{The transfer-based attack success rates (\%) on DGCNN by the proposed ITA attack with our \red{adversarial} transformation model. * indicates the black-box victim model adopts the defense method.}
    \renewcommand{\arraystretch}{1.2}{
    \setlength{\tabcolsep}{1.4mm}{
    \begin{tabular}{|c|c|ccc|}
    \hline
    White-box & Attack & \multicolumn{3}{c|}{Defense: Ours} \\ \cline{3-5}
    Target Model & Methods & PointNet* & PointNet++* & DGCNN \\ \hline
    \multirow{4}*{DGCNN} & FGSM & \red{2.82} & \red{5.93} & \red{11.71}  \\
    ~ & 3D-ADV & \red{5.61} & \red{8.26} & \red{16.38} \\
    ~ & GeoA & \red{9.35} & \red{12.19} & \red{21.07} \\
    ~ & Ours & \red{12.04} & \red{14.97} & \red{25.24} \\ \hline \hline
     White-box & Attack & \multicolumn{3}{c|}{\red{Defense: Vanilla adversarial training \cite{liu2019extending}}} \\ \cline{3-5}
    Target Model & Methods & PointNet* & PointNet++* & DGCNN \\ \hline
    \multirow{4}*{DGCNN} & FGSM & \red{21.72} & \red{25.90} & \red{40.31}  \\
    ~ & 3D-ADV & \red{24.59} & \red{29.11} & \red{45.75}  \\
    ~ & GeoA & \red{29.64} & \red{36.25} & \red{51.26} \\
    ~ & Ours & \red{35.38} & \red{41.79} & \red{56.14}  \\ \hline
    \end{tabular}}}
    \label{tab:defense3}
\end{table}

From these tables, we have \red{three} major observations:
\red{(1) While the vanilla adversarial training strategy (Defense: Vanilla adversarial training \cite{liu2019extending}) and the adversarial training in \cite{sun2020adversarial} (Defense: Adversarial training with \cite{sun2020adversarial}) defend existing attacks in the white-box setting to some extent, our proposed loss in the latent space achieves a significant defense improvement over both methods, thus validating the robustness of our defense model on white-box attacks.}
(2) The transfer-based attack success rates are much lower than the corresponding values in Table \ref{tab:transfer3}, which validates the robustness of the black-box victim model trained by our adversarial training based defense strategy.
(3) In each table, the success rates on the black-box model with our \red{adversarial training} defense are much lower than \red{those of both the vanilla adversarial training strategy and the one in \cite{sun2020adversarial}}. 
This indicates that both intra- and inter-class constraint losses in the latent space are essential for learning invariant information of the clean and corresponding adversarial point clouds.

To further investigate the interpretability of our proposed constraint function in the latent space, we provide some visualization examples of the extracted global features of the clean and adversarial point clouds in Figure \ref{fig:feature}. 
For ease of visualization, we employ Principal Component Analysis (PCA) to reduce the dimensions of global features of each point cloud into three.
As shown in the figure, each column demonstrates the distribution changes of the same feature space of the same point cloud group. Each sub-figure in the first row represents the feature distribution of 75 point cloud instances of 3 object classes (25 instances for each object class). The second and third rows represent the feature distribution of 25 point cloud instances of the adversarial examples without and with our constraint loss function.
We see that, existing adversarial attacks cause ambiguous features locating on the target cloud cluster, while our constraint function in the latent space effectively learns more discriminative and accurate features that pull the ambiguity back to the clean clusters, thus leading to more robust classification results.
\section{Conclusion}
In this paper, we propose a novel Imperceptible Transfer Attack (ITA) to generate qualified adversarial point clouds with high imperceptibility and transferability. 
In particular, instead of perturbing points along xyz directions, we shift points along the normal direction within a strictly bounded width, which alleviates the issue of uneven local distribution and outliers.  
To further enhance the transferability of adversarial examples, we develop an adversarial transformation model to generate the most harmful transformations for structural deformations while enforcing the adversarial examples to resist such distortions, so that the adversarial examples are able to survive under weaker distortions when transferred to attack an unknown black-box model. 
Moreover, to provide a generic defense method for such imperceptible and transferable attacks, we extend the adversarial training based defense approach by introducing additional similarity constraint functions in the latent feature space, which enforces the representations of intra-class point clouds to be closer and those of inter-class point clouds to be discriminative.
Experimental results on \red{five} popular classification models (PointNet, PointNet++, DGCNN, \red{PointTrans. and PointMLP}) validate the imperceptibility and transferability of the proposed ITA attack when compared with state-of-the-art methods, as well as the robustness under noise perturbation or incomplete data.   
We also verify the superiority of our defense method and provide some interpretation via visualization of the feature space.

\ifCLASSOPTIONcaptionsoff
  \newpage
\fi



\bibliographystyle{IEEEtran}
\bibliography{reference.bib}

\end{document}